\documentclass[preprint,12pt]{RAS_template/elsarticle}

\usepackage{amsmath}
\usepackage{amssymb}
\usepackage{amsfonts}
\usepackage{graphicx}
\usepackage{subcaption}
\usepackage{float}
\usepackage{placeins}
\usepackage{textcomp}
\usepackage{xcolor}
\usepackage{caption}
\usepackage{cuted}

\usepackage[hidelinks]{hyperref}

\journal{Robotics and Autonomous Systems}

\begin{document}

\begin{frontmatter}

\title{
FORTE: Task-Adaptive Force Capability Optimization for Mobile Manipulators
}

\author[hri2,XJ]{Xiao Wang}
\ead{xiao.wang@iit.it}
\author[hri2,drim]{Heng Zhang}
\author[hri2]{Gokhan Solak}
\author[XJ]{Fei Zhao\corref{cor1}}
\ead{ztzhao@mail.xjtu.edu.cn}
\cortext[cor1]{Corresponding author.}
\author[hri2]{Arash Ajoudani}

\affiliation[hri2]{
        organization={HRI$^2$ Lab, Istituto Italiano di Tecnologia},
        city={Genoa},
        postcode={16163},
        country={Italy}
        }
\affiliation[XJ]{
organization={School of Mechanical Engineering, Xi'an Jiaotong University},
city={Xi'an},
postcode={710049},
country={China}
}
\affiliation[drim]{organization={Ph.D. program of national interest in Robotics and Intelligent Machines (DRIM) and University of Genova},
        city={Genoa},
        postcode={16126},
        country={Italy}}

\begin{abstract}
Effective physical interaction control in robotic manipulation requires not only kinematically feasible motion but also sufficient force-interaction capability. Existing redundancy resolution methods often ignore task-specific force demands or maximize the force capability indiscriminately, sacrificing dexterity when large force margins are unnecessary. We propose a task-oriented force capability optimization framework for redundant mobile manipulators. A Vision-Language Model (VLM) infers object physical properties from an RGB image and a task description, generating a desired task-force sequence that captures gravitational and inertial demands. We then define a task-oriented force capability metric as the signed distance between a task-force uncertainty ball and the dynamic residual force polytope (RFP), quantifying compatibility between task demands and the robot's remaining actuation capacity. This metric is incorporated, alongside manipulability, joint-limit avoidance, trajectory smoothness, and base-oscillation suppression, into a whole-body multi-objective trajectory-optimization problem. Experiments on a mobile manipulator performing lifting and single-point-holding tasks under varying payload conditions demonstrate that the proposed method provides sufficient force capability for heavy loads while preserving high manipulability for light loads. This yields a task-adaptive balance that fixed capability-maximizing baselines (RFP inscribed radius, RFP cone) and manipulability-only optimization fail to achieve. The core implementation is publicly available at \url{https://github.com/yeying256/FORTE}.

\end{abstract}


\end{frontmatter}

\section{Introduction}

Successful manipulation requires not only accurate motion execution but also sufficient physical interaction capability~\cite{tsuji2025survey}. Although a desired end-effector trajectory may be kinematically feasible, task completion can still fail if the robot is unable to generate the forces required to interact with the environment. This issue becomes particularly important in tasks involving heavy payload transportation, object pushing, drawer opening, and other contact-rich interactions, where the required interaction forces may vary significantly throughout task execution~\cite{zhang2025safe,Where2Act,VoxPoser,RT-2}.

\begin{figure}[!htbp]
    \centering
    \includegraphics[width=0.95\linewidth]{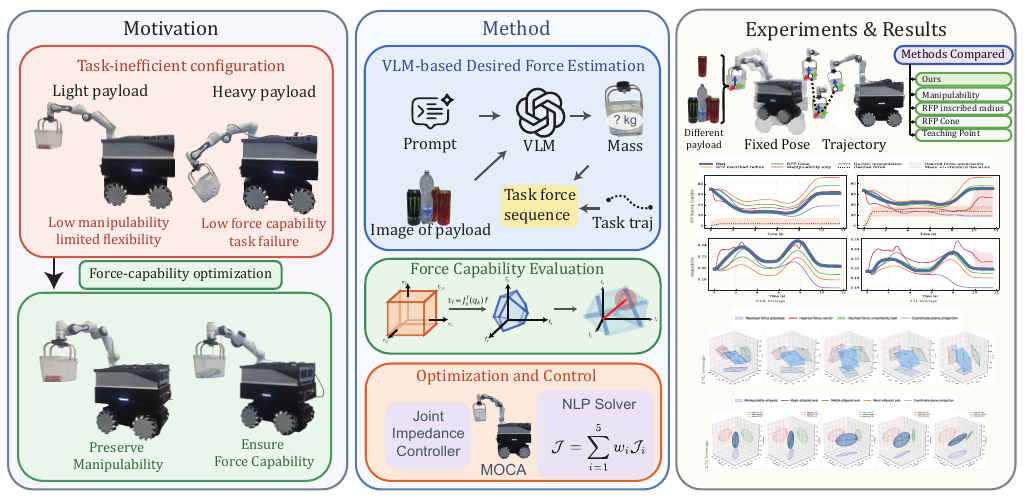}
    \caption{
Overview of the proposed task-oriented force-capability optimization framework.
Task-inefficient configurations may cause insufficient force capability for heavy payloads or limited manipulability for light payloads.
The proposed framework estimates the task-force alignment, evaluates its compatibility with the robot's residual force capability, and optimizes the whole-body configuration to achieve task-sufficient force capability while preserving manipulability.
    }
    \label{fig:title_overview}
\end{figure}

Robotic systems possess well-defined actuation limits, which restrict the set of achievable end-effector forces~\cite{DynamicManipulabilityRobot1985,chiacchioForcePolytopeForce1997}. Furthermore, uncertainties arising from external disturbances, payload variations, and modeling errors may further increase the force demand during task execution. Consequently, ensuring that the robot maintains sufficient force-generation capability along the planned trajectory is critical for robust task completion~\cite{ferrolhoResidualForcePolytope2021a}.

For redundant robotic systems such as mobile manipulators, infinitely many joint configurations can satisfy the same end-effector pose constraint. Although these configurations achieve identical task-space objectives, their kinematic and dynamic properties may differ significantly due to variations in Jacobian mappings, joint torque distributions, and proximity to joint limits. In particular, the force-generation capability associated with different configurations can vary considerably. Therefore, redundancy can be exploited to reshape the robot's end-effector capability while preserving the desired end-effector trajectory, thereby improving its ability to satisfy task requirements and reject external disturbances~\cite{Ligeois1977AutomaticSC,sicilianoGeneralFrameworkManaging1991,bassoTaskPriorityControlRedundant2020,sentisWholebodyControlFramework2006a}.

Extensive research has examined redundancy resolution to improve robotic performance. Manipulability-based methods optimize robot configurations to enhance velocity transmission capability and dexterity~\cite{Yoshikawa1985Manipulability,Dietrich2015}. At the same time, other studies employ force manipulability ellipsoids, force polytopes, and related metrics to evaluate and maximize force-generation capability~\cite{DynamicManipulabilityRobot1985,chiacchioForcePolytopeForce1997,gargWrenchCapabilityAnalysis2009,ferrolhoResidualForcePolytope2021a}. However, most existing approaches focus on improving capability metrics themselves, while the relationship between task force requirements and robot capability has received limited attention. In practice, a larger force-generation capability does not necessarily imply better task performance; reliable task execution requires the robot's available capability to satisfy the task demand adequately.

Moreover, existing force capability optimization methods typically assume that task requirements are known \emph{a priori}. In practical scenarios, however, robots often need to infer task force requirements from environmental observations and object properties. Recent vision-language-based manipulation frameworks have demonstrated the ability to infer semantic task constraints and interaction strategies directly from visual observations~\cite{RT-2,VoxPoser,Where2Act,zhang2025omnivic}. Overall, existing force-capability methods are not task-aware. Therefore, estimating task force requirements from perceptual information and exploiting robot redundancy for task-aware configuration optimization remain open challenges~\cite{PhysX-Anything,Where2Act,VoxPoser,RT-2}.

To address these challenges, this paper proposes a task-oriented capability optimization framework for redundant mobile manipulators, as illustrated in Fig.~\ref{fig:title_overview}. Instead of directly optimizing capability metrics, the proposed framework first estimates task force requirements from perceptual information. A task-oriented force margin metric based on the residual force polytope is then developed to quantify the compatibility between task demands and robot capabilities. Finally, the proposed metric is integrated with manipulability, joint-limit avoidance, trajectory smoothness, and other performance objectives within a multi-objective optimization framework, enabling the robot to adaptively adjust its configuration according to task requirements while balancing force capability and motion performance.

The main contributions of this work are summarized as follows:
\begin{itemize}

\item \textbf{We propose FORTE, a task-oriented whole-body redundancy optimization framework for mobile manipulators.} The framework jointly optimizes force capability, manipulability, joint-limit avoidance, trajectory smoothness, and base motion while preserving the desired end-effector trajectory.

\item We develop a VLM-based task-force estimation method that infers task-force requirements directly from an RGB image, a task description, and the desired trajectory, enabling perception-driven whole-body optimization.

\item We introduce a task-oriented force-capability metric based on the residual force polytope. Extensive experiments on the MOCA mobile manipulator demonstrate the effectiveness of the proposed framework.

\end{itemize}

\section{Related Work}

\subsection{Task Requirement Estimation for Robotic Manipulation}

Estimating task requirements is essential for robotic manipulation, especially when the robot must interact with objects whose physical properties are unknown. Traditional approaches rely on analytical models and dynamic identification. Atkeson~\cite{atkesonEstimationInertialParameters1986} estimated inertial parameters of manipulators and payloads for model-based force and torque computation, while Swevers~\cite{sweversOptimalRobotExcitation1997} systematically studied robot dynamic identification for obtaining mass, inertia, and friction parameters. To handle unknown payloads during execution, Kubus~\cite{kubusOnlineRigidObject2007} estimated object inertial properties online and used them to improve manipulation performance.

Recent vision-based and vision-language methods further enable robots to infer physical and semantic task information from perceptual observations. Image-based methods have been explored for estimating object mass and physical properties~\cite{image2mass,PhysX-Anything}. Meanwhile, works such as Where2Act~\cite{Where2Act}, VoxPoser~\cite{VoxPoser}, and RT-2~\cite{RT-2} demonstrate that visual and vision-language representations can support interaction affordance prediction, task-constraint generation, and semantic robotic decision-making.

However, most existing methods focus on estimating object properties or generating task plans, while the estimated information is rarely converted into explicit task force requirements for robot capability analysis and redundancy optimization. In contrast, this work utilizes perceptual information to estimate task force requirements and incorporates them into a task-oriented force capability optimization framework.

\subsection{Force Capability Analysis of Redundant Robots}

Redundant robots can exploit additional degrees of freedom to improve motion and force generation performance. A classical measure is manipulability, introduced by Yoshikawa~\cite{Yoshikawa1985Manipulability}, which quantifies velocity transmission capability and has been widely used for singularity avoidance and redundancy resolution~\cite{Dietrich2015}. Yoshikawa later extended this idea to force manipulability~\cite{DynamicManipulabilityRobot1985}, describing achievable end-effector forces under joint torque constraints.

Compared with ellipsoid-based measures, force polytopes provide a more accurate representation of achievable force sets because they explicitly incorporate actuator limits. Chiacchio~\cite{chiacchioForcePolytopeForce1997} formulated force polytopes for redundant manipulators, and Garg~\cite{gargWrenchCapabilityAnalysis2009} extended related ideas to wrench capability analysis. More recently, Ferrolho~\cite{ferrolhoResidualForcePolytope2021a} introduced the residual force polytope, which characterizes the remaining task-space force capability after accounting for gravity compensation and dynamic constraints.

Although these methods provide useful tools for evaluating or maximizing robot force capability, they are generally not task-aware. A larger force capability does not necessarily lead to better task performance. For many manipulation tasks, once the available force capability is sufficient to satisfy the task requirement, further increasing it may provide limited benefit while sacrificing manipulability, dexterity, or motion quality. Therefore, the key problem is not to maximize force capability indiscriminately, but to evaluate whether the robot has sufficient force capability for the specific task. This work addresses this gap by explicitly measuring the compatibility between task force requirements and the robot's residual force capability.

\subsection{Redundancy Resolution and Multi-Objective Optimization}

Redundancy resolution aims to exploit additional degrees of freedom while satisfying primary task constraints. Li{\'e}geois~\cite{Ligeois1977AutomaticSC} introduced the gradient projection method for optimizing secondary objectives in the null space of the primary task. Task-priority and null-space projection frameworks were further developed by Basso~\cite{bassoTaskPriorityControlRedundant2020} and Siciliano~\cite{sicilianoGeneralFrameworkManaging1991} for hierarchical task execution. In these frameworks, objectives such as manipulability maximization, joint-limit avoidance, energy reduction, and trajectory smoothness are commonly used to improve robot performance.

For mobile manipulators, redundancy resolution becomes more important because the mobile base and manipulator arm must be coordinated. Whole-body control and optimization frameworks, such as that of Sentis~\cite{sentisWholebodyControlFramework2006a}, exploit the combined degrees of freedom of the system under task and physical constraints.

Nevertheless, most existing redundancy optimization methods rely on predefined objectives that are largely independent of the physical task requirements. Capability-oriented approaches often favor configurations with larger force capability, even when such capability is unnecessary for the current task. This may lead to overly conservative configurations and reduced motion quality. In contrast, this work incorporates estimated task force requirements into redundancy optimization. The proposed framework first estimates task force requirements from perceptual information, then evaluates their compatibility with the robot's residual force capability, and finally performs multi-objective whole-body optimization to achieve task-sufficient force capability while preserving manipulability, joint safety, and motion smoothness.

\section{Preliminaries}
\label{sec:robotics_background}
This section briefly introduces the robotics concepts used throughout the remainder of this paper. First, the whole-body joint impedance control framework employed for executing the optimized joint-space trajectories on the mobile manipulator is presented. Then, the residual force polytope (RFP) is introduced as a geometric representation of the robot's remaining force generation capability under actuator constraints and dynamic loading conditions. These concepts provide the foundation for the task-oriented force capability analysis and trajectory optimization framework developed in the following section.

\subsection{Joint Impedance Control of Mobile Manipulator}

Since the proposed optimization framework directly generates whole-body joint-space trajectories, a joint impedance controller is employed to track the optimized configurations during execution. Compared with task-space controllers, joint impedance control allows the optimized trajectories to be executed directly while preserving compliant interaction behavior.

For an $n$-DoF robotic system, the desired joint-space impedance behavior is described by
\begin{equation}
M_d\ddot e+D_d\dot e+K_de = - \tau_{ext},
\end{equation}
where $e=q_d-q$ is the joint tracking error, and $q_d,q\in\mathbb{R}^{n}$ are the desired and measured joint positions, respectively. $M_d,D_d,K_d\in\mathbb{R}^{n\times n}$ are the desired inertia, damping, and stiffness matrices, and $\tau_{\mathrm{ext}}\in\mathbb{R}^{n}$ denotes the external force acting on the robot.

For the MOCA mobile manipulator~\cite{yuqiang_imp}, the whole-body generalized coordinate vector is defined as
$q=[x_b,y_b,\theta_b,q_{a,1},\cdots,q_{a,7}]^{T}\in\mathbb{R}^{10}$,
where $(x_b,y_b)$ and $\theta_b$ denote the planar position and yaw angle of the mobile base, respectively, and $q_a\in\mathbb{R}^{7}$ represents the Franka arm configuration.

The mobile base provides a velocity control interface, while the manipulator provides a torque control interface. Accordingly, different low-level controllers are adopted for the two subsystems.

For the manipulator, the desired inertia is chosen to match the arm inertia, i.e., $M_{d,a}=M_a(q_a)$.
With the joint tracking error defined as $e_a=q_{a,d}-q_a$, the commanded joint torque is computed as
\begin{equation}
\tau_a
=
M_a(q_a)\ddot q_{a,d}
+
C_a(q_a,\dot q_a)
+
g(q_a)
+
K_ae_a
+
D_a\dot e_a
+
\tau_F,
\end{equation}
where $\dot e_a=\dot q_{a,d}-\dot q_a$, $M_a(q_a)$ is the arm inertia matrix, and $K_a$ and $D_a$ are the joint stiffness and damping matrices, respectively.
Here, $C_a(q_a,\dot q_a)$ denotes the Coriolis and centrifugal compensation term, $g(q_a)$ denotes the gravity compensation term, and $\tau_F$ is the feedforward torque generated from the desired task-space force.

To guarantee safe operation, the commanded torque is saturated according to the actuator limits:
\begin{equation}
\tau_{a,i}
\leftarrow
\mathrm{sat}
\left(
\tau_{a,i},
-\tau_{i,\max},
\tau_{i,\max}
\right),
\qquad
i=1,\ldots,7.
\end{equation}

For the mobile base, the generalized coordinate is $q_b=[x_b,y_b,\theta_b]^T$, with tracking error $e_b=q_{b,d}-q_b$. The corresponding virtual generalized force $\tau_b=[F_x,F_y,\tau_z]^T$ is computed as
\begin{equation}
\tau_b
=
K_be_b
-
D_b\dot q_b,
\end{equation}
and is converted into the commanded base velocity $v_b=[v_x,v_y,\omega_z]^T$ through the admittance model
\begin{equation}
M_b\dot v_b+B_bv_b=\tau_b.
\end{equation}

Consequently, the optimizer outputs the desired whole-body trajectory $q_d=[q_{b,d}^{T},q_{a,d}^{T}]^{T}$, which is executed through torque-based joint impedance control for the manipulator and velocity-based admittance control for the mobile base.

Although the optimizer generates trajectories for the entire mobile manipulator, the force capability analysis presented in this work is performed only for the manipulator arm. Specifically, the mobile base is regarded as a kinematic redundancy source that changes the arm configuration and therefore indirectly influences the arm Jacobian and residual force polytope. The residual force polytope is computed solely from the arm actuator torque limits, while the mobile base is assumed to accurately execute the planned motion through the low-level admittance controller.

\subsection{Residual Force Polytope}

\begin{figure}[!htbp]
    \centering
    \includegraphics[width=0.95\linewidth]{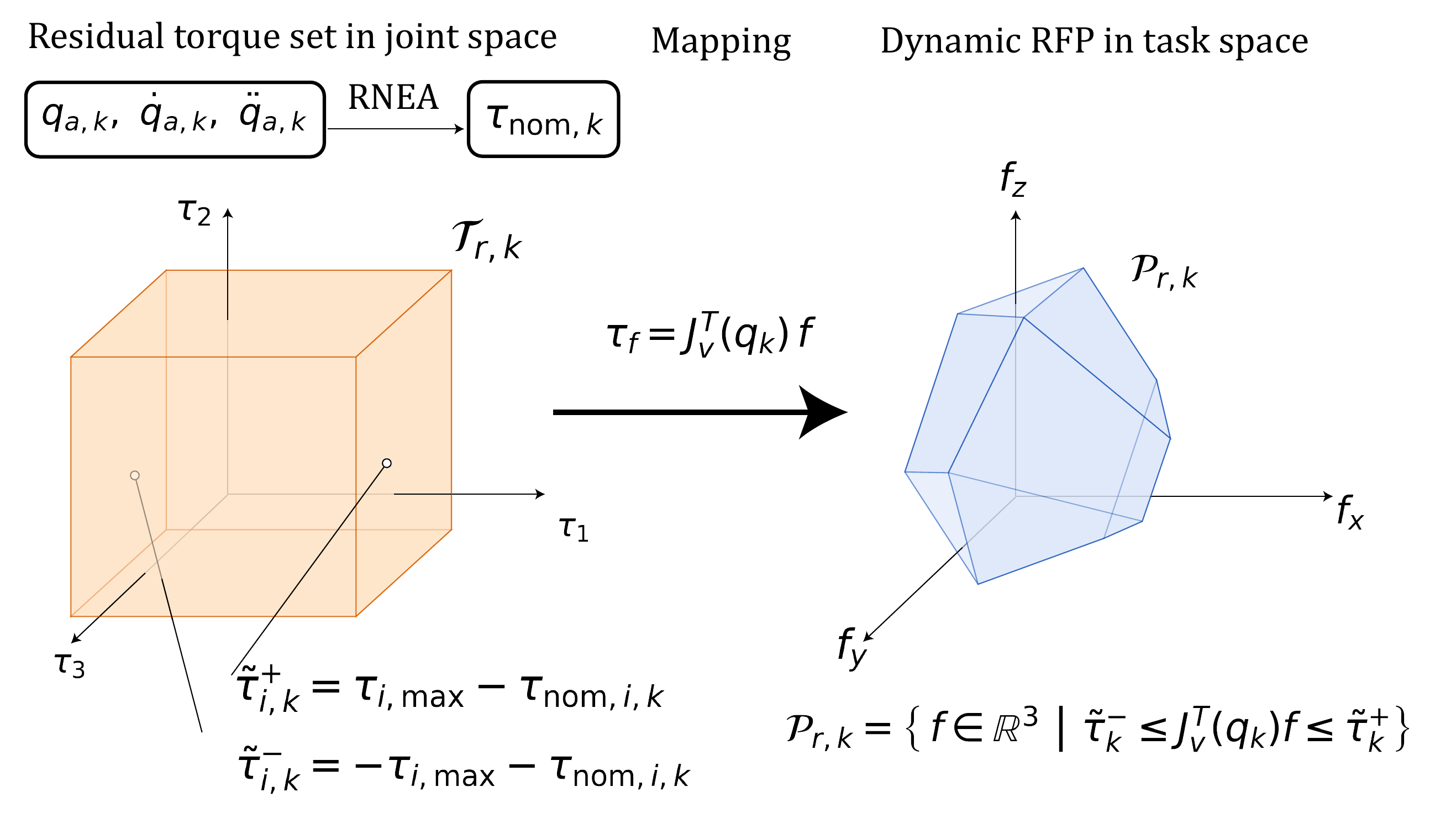}
    \caption{
    Illustration of the construction of the dynamic Residual Force Polytope (RFP). 
    The nominal torque $\tau_{\mathrm{nom},k}$ required to execute the trajectory is first computed from the joint position, velocity, and acceleration using inverse dynamics. 
    The residual torque margins define the remaining admissible torque set $\mathcal{T}_{r,k}$ in joint space. 
    By substituting the force-to-torque mapping $\tau_f=J_v^{T}(q_k)f$ into the residual torque constraints, the joint-space residual actuation capability is mapped into the task-space dynamic residual force polytope $\mathcal{P}_{r,k}$.
    }
    \label{fig:rfp_mapping}
\end{figure}

To evaluate the force generation capability that remains available during trajectory execution, this work adopts the Residual Force Polytope (RFP)~\cite{ferrolhoResidualForcePolytope2021a} as a task-space representation of the robot's remaining actuation capability. Unlike conventional force polytopes that only consider actuator limits, the RFP explicitly accounts for the torque required to execute the nominal trajectory. Consequently, it characterizes the set of end-effector forces that can still be generated while simultaneously tracking the desired motion. Fig.~\ref{fig:rfp_mapping} illustrates the construction of the dynamic residual force polytope~\cite{chengConstrainedDynamicForce2026} from the residual joint torque set.

In this work, the residual force polytope is defined only for the 7-DoF manipulator. Although the mobile base participates in trajectory optimization, its wheel actuation limits are not explicitly incorporated into the force capability analysis. Instead, the optimized base motion modifies the manipulator configuration, which in turn changes the Jacobian and the nominal arm torque required for trajectory execution. Consequently, the proposed framework evaluates the residual force capability of the arm conditioned on the planned whole-body configuration.

According to the principle of virtual work, the end-effector force
$f\in\mathbb{R}^{3}$ and the corresponding joint torque vector satisfy

\begin{equation}
\tau_f
=
J_v^{T}(q)f,
\end{equation}

where $J_v(q)\in\mathbb{R}^{3\times 7}$ denotes the translational Jacobian of the manipulator.

The manipulator joints are subject to actuator torque limits. Let
$\tau_{i,\max}$ denote the maximum allowable torque of the $i$-th joint. During trajectory execution, part of the available actuator capability must be consumed to generate the desired motion. Therefore, the actual force generation capability depends on the remaining torque margins after accounting for the nominal trajectory dynamics.

For a discrete trajectory, the joint velocity and acceleration are approximated by finite differences,
$\dot q_{a,k}=(q_{a,k}-q_{a,k-1})/\Delta t_k$ and
$\ddot q_{a,k}=(\dot q_{a,k}-\dot q_{a,k-1})/\Delta t_k$,
where $\Delta t_k=t_k-t_{k-1}$.

Given the joint position $q_{a,k}$, velocity $\dot q_{a,k}$, and acceleration $\ddot q_{a,k}$, the nominal torque required to execute the trajectory is computed using inverse dynamics:

\begin{equation}
\tau_{\mathrm{nom},k}
=
\mathrm{RNEA}
\left(
q_{a,k},
\dot q_{a,k},
\ddot q_{a,k}
\right),
\end{equation}

where RNEA denotes the Recursive Newton-Euler Algorithm.

The nominal trajectory torque occupies part of the available actuator capability. Therefore, the remaining torque margins are defined as the distance between the actuator torque limits and the nominal torque required for trajectory execution.

Specifically, the positive and negative residual torque margins are
$\tilde\tau_{i,k}^{+}=\tau_{i,\max}-\tau_{nom,i,k}$
and
$\tilde\tau_{i,k}^{-}=-\tau_{i,\max}-\tau_{nom,i,k}$,
respectively.
$\tilde{\tau}_{i,k}^{+}$ and
$\tilde{\tau}_{i,k}^{-}$
represent the remaining actuator capability available for generating additional end-effector forces in the positive and negative torque directions, respectively.

Accordingly, the residual joint torque set at the $k$-th waypoint is defined as

\begin{equation}
\mathcal{T}_{r,k}
=
\left\{
\tau_f \in \mathbb{R}^{7}
\;\middle|\;
\tilde{\tau}_{i,k}^{-}
\le
\tau_{f,i}
\le
\tilde{\tau}_{i,k}^{+},
\quad
i=1,\ldots,7
\right\}.
\end{equation}

Substituting the virtual-work relation into the residual torque constraints yields the dynamic residual force polytope

\begin{equation}
\mathcal{P}_{r,k}
=
\left\{
f\in\mathbb{R}^{3}
\;\middle|\;
\tilde{\tau}_{k}^{-}
\le
J_v^{T}(q_k)f
\le
\tilde{\tau}_{k}^{+}
\right\},
\end{equation}

where
$\tilde\tau_k^{+}=[\tilde\tau_{1,k}^{+},\cdots,\tilde\tau_{7,k}^{+}]^T$
and
$\tilde\tau_k^{-}=[\tilde\tau_{1,k}^{-},\cdots,\tilde\tau_{7,k}^{-}]^T$. The polytope can be further expressed in half-space form as

\begin{equation}
\mathcal{P}_{r,k}
=
\left\{
f\in\mathbb{R}^{3}
\mid
A_k f
\le
b_k
\right\},
\end{equation}

where
$A_k=[J_v(q_k)\; -J_v(q_k)]^{T}\in\mathbb{R}^{14\times3}$
and
$b_k=[(\tilde{\tau}_{k}^{+})^{T}\; -(\tilde{\tau}_{k}^{-})^{T}]^{T}
\in\mathbb{R}^{14}$.
The resulting dynamic residual force polytope represents the set of admissible end-effector forces that can be generated while simultaneously executing the nominal trajectory. Since the nominal torque depends on the robot configuration, velocity, and acceleration, the polytope's shape and size vary along the trajectory. Consequently, even when a robot exhibits strong static force capability at a given configuration, its remaining force generation capability may decrease significantly during high-speed or high-acceleration motions. Therefore, the dynamic residual force polytope provides a more realistic description of the force capability available during task execution. The proposed formulation assumes accurate base motion and neglects wheel actuation and stability constraints. Therefore, the residual force polytope represents the arm's residual force capability under the planned whole-body configuration.

\section{Method}

\begin{figure*}[t]
    \centering
    \includegraphics[width=0.95\textwidth]{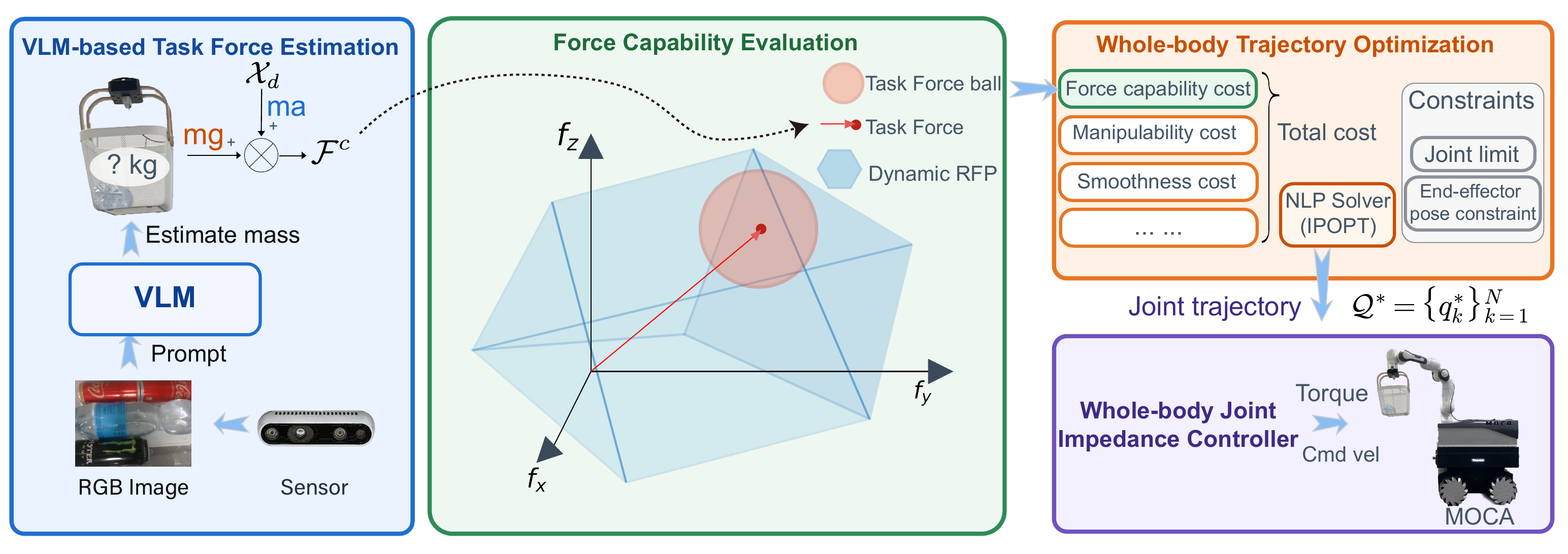}
    \caption{
    Overview of the proposed task-oriented force capability optimization framework.
    The framework consists of three main modules: VLM-based desired force estimation, force capability evaluation, and whole-body trajectory optimization.
    Given a task description $\mathcal{D}$, an RGB image $\mathcal{I}$, and a desired end-effector trajectory $\mathcal{X}_d$, the VLM estimates task-related physical parameters and generates the desired force sequence $\mathcal{F}^{c}$.
    The desired force at each trajectory point is modeled as the center of a task force uncertainty ball, while the Dynamic Residual Force Polytope (Dynamic RFP) characterizes the remaining force generation capability under the current robot configuration.
    The minimum signed distance between the task force uncertainty ball and the Dynamic RFP is used to construct the force capability cost.
    Finally, the force capability cost is jointly optimized with manipulability, joint-limit avoidance, trajectory smoothness, and base oscillation suppression using a constrained nonlinear programming solver to obtain the optimal whole-body trajectory.
    }
    \label{fig:method_overview}
\end{figure*}

The overall framework of the proposed method is illustrated in Fig.~\ref{fig:method_overview}. Given a task description $\mathcal{D}$, an environment image $\mathcal{I}$, and a desired end-effector trajectory $\mathcal{X}_d$, the objective is to determine a whole-body configuration trajectory that satisfies the task force requirements while maintaining favorable kinematic and dynamic properties.

First, a VLM is employed to infer task-relevant physical properties from $\mathcal{D}$ and $\mathcal{I}$. For lifting tasks, the estimated object properties are combined with the desired trajectory to generate a desired force sequence $\mathcal{F}^{c}$, which characterizes the force requirements throughout task execution.

Second, the desired force sequence is incorporated into a force capability evaluation module. At each trajectory point, the desired force is modeled as the center of a task force uncertainty ball. By combining the translational Jacobian and the residual joint torque margins, a Dynamic Residual Force Polytope (Dynamic RFP) is constructed. The minimum signed distance between the task force ball and the Dynamic RFP is then used to quantify the compatibility between the task force requirements and the robot's available force generation capability.

Finally, the proposed force capability metric is integrated into a whole-body multi-objective trajectory optimization framework together with manipulability, joint-limit avoidance, trajectory smoothness, and base oscillation suppression objectives. The resulting nonlinear programming (NLP) problem is solved under end-effector trajectory constraints to obtain the optimal whole-body configuration sequence

\begin{equation}
\mathcal{Q}^{*}
=
\left\{
q_k^{*}
\right\}_{k=1}^{N}.
\end{equation}

The optimized trajectory is subsequently executed by the whole-body joint impedance controller introduced in Section~\ref{sec:robotics_background}.

\subsection{VLM-based Desired Force Estimation}

\begin{figure}[!htbp]
    \centering
    \includegraphics[width=0.98\linewidth]{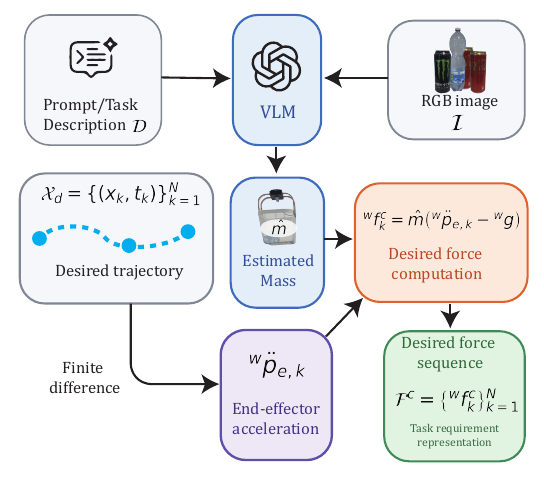}
    \caption{
    Overview of the proposed desired force estimation module. 
    Given a prompt-based task description $\mathcal{D}$ and an RGB image $\mathcal{I}$, the VLM estimates the mass $\hat{m}$ of the manipulated object. 
    Meanwhile, the desired end-effector trajectory $\mathcal{X}_d$ is processed by finite differences to obtain the end-effector acceleration ${}^{w}\ddot{p}_{e,k}$. 
    The estimated mass and the trajectory-induced acceleration are then combined with gravity compensation to compute the desired task force ${}^{w}f_k^{c}=\hat{m}({}^{w}\ddot{p}_{e,k}-{}^{w}g)$ at each waypoint. 
    The resulting desired force sequence $\mathcal{F}^{c}$ serves as the task requirement representation for the subsequent force capability evaluation and trajectory optimization.
    }
    \label{fig:vlm_desired_force_estimation}
\end{figure}

This work employs a VLM to infer task-relevant physical properties from a task description $\mathcal{D}$ and an environment image $\mathcal{I}$, and subsequently generates the desired force required during task execution. Fig.~\ref{fig:vlm_desired_force_estimation} illustrates the pipeline of the proposed VLM-based desired force estimation module.

For lifting tasks, the VLM first estimates the mass of the manipulated object as

\begin{equation}
\hat{m}
=
\Phi_{\mathrm{VLM}}
\left(
\mathcal{I},
\mathcal{D}
\right),
\end{equation}

where $\Phi_{\mathrm{VLM}}(\cdot)$ denotes the VLM inference process and $\hat{m}$ is the estimated object mass.

Given the desired end-effector trajectory

\begin{equation}
\mathcal{X}_d
=
\left\{
(x_k,t_k)
\right\}_{k=1}^{N},
\qquad
x_k \in SE(3),
\end{equation}

the desired force at the $k$-th trajectory waypoint is defined as

\begin{equation}
{}^{w}f_{k}^{c}
=
\hat{m}
\left(
{}^{w}\ddot{p}_{e,k}
-
{}^{w}g
\right),
\end{equation}

where ${}^{w}\ddot{p}_{e,k}$ denotes the end-effector acceleration obtained from finite differences of the discrete trajectory, and ${}^{w}g
=
\begin{bmatrix}
0 &
0 &
-9.81
\end{bmatrix}^{T}
\ \mathrm{m/s^2}$
is the gravitational acceleration vector expressed in the world frame.

The desired force consists of both the inertial force induced by trajectory motion and the force required to compensate for object gravity. In this work, the desired force is defined as the force that must be generated by the manipulator to support and transport the object along the desired trajectory. Therefore, under static conditions (${}^{w}\ddot{p}_{e,k}=0$), the desired force reduces to the upward gravity-compensation force.

Lifting tasks can be regarded as a special case of force application tasks. During lifting, the robot must continuously generate forces to compensate for both object weight and motion-induced inertial effects. Therefore, the desired force vector simultaneously determines the required force magnitude and the task force direction.

The resulting desired force sequence is

\begin{equation}
\mathcal{F}^{c}
=
\left\{
{}^{w}f_k^{c}
\right\}_{k=1}^{N}.
\end{equation}

The desired force sequence serves as the task requirement representation throughout the remainder of the framework. It is subsequently used to construct the task force uncertainty ball and to evaluate the compatibility between task requirements and the robot's available force generation capability.

\subsection{Force Capability Metric Based on Residual Force Polytope}

\begin{figure}[!htbp]
    \centering

    \begin{subfigure}[t]{0.32\linewidth}
        \centering
        \includegraphics[width=\linewidth]{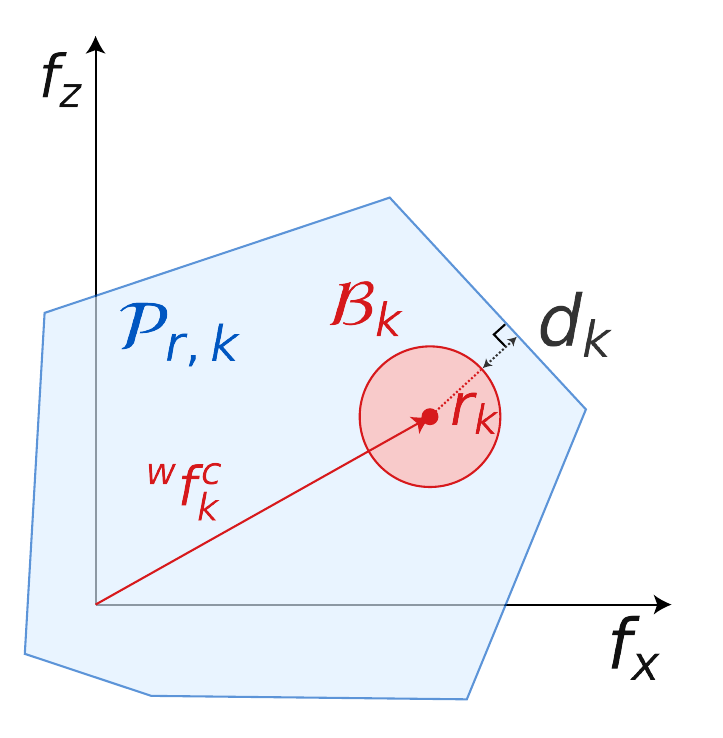}
        \caption{$d_k>0$: inside}
        \label{fig:force_metric_inside}
    \end{subfigure}
    \hfill
    \begin{subfigure}[t]{0.32\linewidth}
        \centering
        \includegraphics[width=\linewidth]{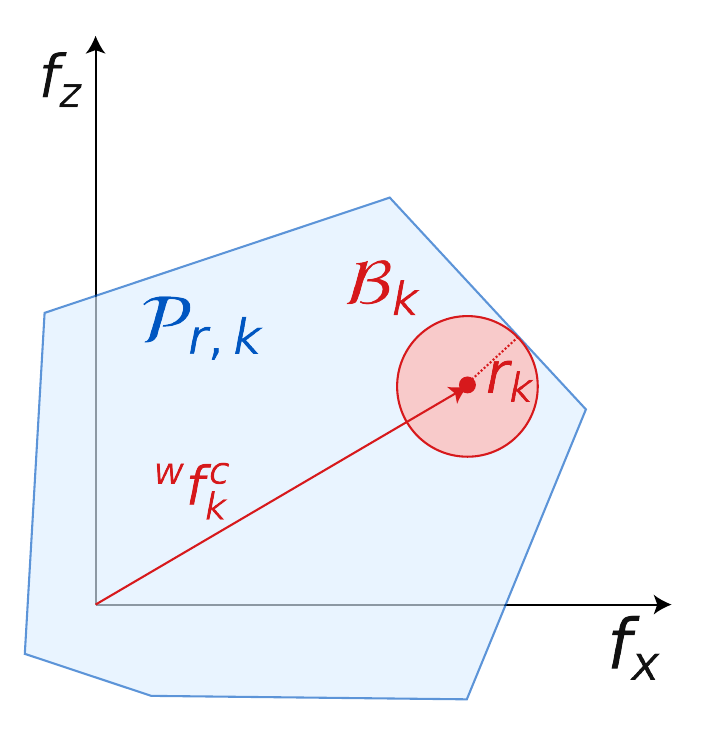}
        \caption{$d_k=0$: tangent}
        \label{fig:force_metric_tangent}
    \end{subfigure}
    \hfill
    \begin{subfigure}[t]{0.32\linewidth}
        \centering
        \includegraphics[width=\linewidth]{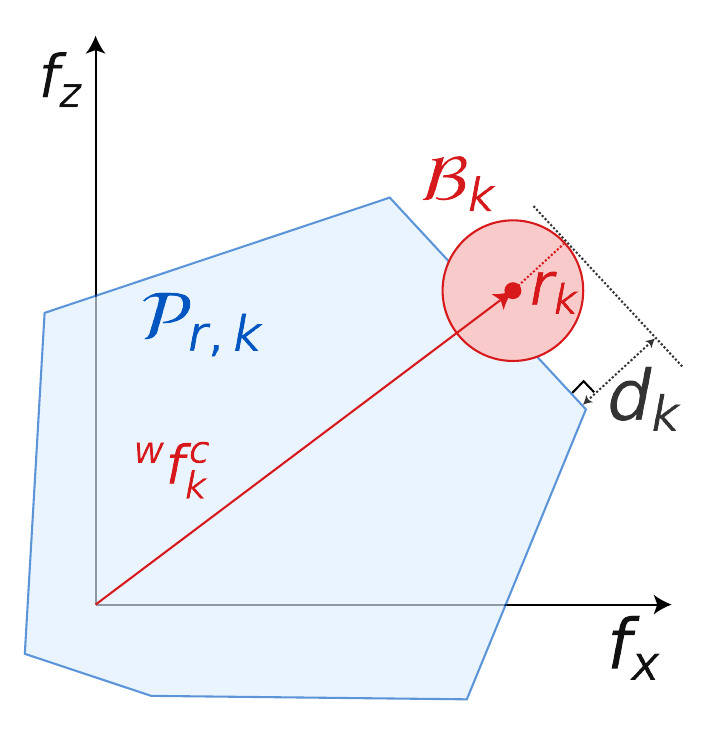}
        \caption{$d_k<0$: outside}
        \label{fig:force_metric_outside}
    \end{subfigure}

    \caption{
    Geometric illustration of the proposed force capability metric based on the relative position between the task force uncertainty ball and the Dynamic Residual Force Polytope (Dynamic RFP). The red circle denotes the task force uncertainty ball $\mathcal{B}_k$, whose center is the desired task force ${}^{w}f_k^{c}$ and whose radius is $r_k$. The blue polygon represents the Dynamic RFP $\mathcal{P}_{r,k}$, which characterizes the available residual force generation capability at the $k$-th waypoint. 
    (a) The task force ball is fully contained inside the Dynamic RFP, yielding a positive force capability margin $d_k>0$. 
    (b) The task force ball is tangent to the boundary of the Dynamic RFP, corresponding to the critical case $d_k=0$. 
    (c) The task force ball extends outside the Dynamic RFP, indicating an insufficient force capability margin with $d_k<0$.
    }
    \label{fig:force_capability_metric}
\end{figure}

The objective of this section is to quantify the compatibility between the task force requirements and the robot's available force generation capability. To this end, a task-oriented force capability metric is constructed based on the relative position between the task force uncertainty set and the Dynamic Residual Force Polytope (Dynamic RFP). Fig.~\ref{fig:force_capability_metric} illustrates the geometric meaning of the signed distance $d_k$ between the task force uncertainty ball and the Dynamic RFP. A positive signed distance indicates that the task-force requirement is satisfied with the prescribed safety margin, whereas a negative signed distance indicates that this safety margin is not fully maintained.

The discrete end-effector trajectory is defined as
\begin{equation}
\mathcal{X}_d
=
\left\{
x_k
\mid
k=1,\ldots,N,\;
x_k \in SE(3)
\right\},
\end{equation}
where $x_k$ denotes the desired end-effector pose at the $k$-th waypoint.

Correspondingly, the desired force sequence generated by the task understanding module is defined as
\begin{equation}
\mathcal{F}^{c}
=
\left\{
{}^{w}f_k^{c}
\mid
k=1,\ldots,N,\;
{}^{w}f_k^{c}\in\mathbb{R}^{3}
\right\},
\end{equation}
where ${}^{w}f_k^{c}$ denotes the desired task force expressed in the world frame and serves as the center of the task force uncertainty set.

To account for uncertainties originating from VLM estimation errors, model mismatch, and unmodeled task variations, the desired task force is represented as a closed ball in the task-force space. At the $k$-th trajectory waypoint, the task force uncertainty set is defined as
\begin{equation}
\mathcal{B}_k
=
\left\{
{}^{w}f \in \mathbb{R}^{3}
\;\middle|\;
\left\|
{}^{w}f
-
{}^{w}f_k^{c}
\right\|_2
\leq
r_k
\right\},
\end{equation}
where $r_k$ denotes the uncertainty radius associated with the desired task force.

In this work, only translational force generation capability is considered. Let the whole-body configuration corresponding to the $k$-th trajectory waypoint be $ q_k
=
\begin{bmatrix}
q_{b,k}^{T}
&
q_{a,k}^{T}
\end{bmatrix}^{T},$ where $q_{b,k}$ denotes the mobile base state and $q_{a,k}\in\mathbb{R}^{n_a}$ denotes the arm joint configuration.

The translational Jacobian expressed in the world frame is
\begin{equation}
{}^{w}J_{t,k}
=
{}^{w}J_t(q_k)
=
\begin{bmatrix}
{}^{w}j_{1,k}
&
\cdots
&
{}^{w}j_{n_a,k}
\end{bmatrix}
\in
\mathbb{R}^{3\times n_a},
\end{equation}
where $n_a$ denotes the number of arm joints and ${}^{w}j_{i,k}\in\mathbb{R}^{3}$ is the $i$-th column of ${}^{w}J_{t,k}$.

Each joint torque constraint induces a corresponding pair of half-space constraints in the task-force space. Therefore, the normal vector associated with the $i$-th joint constraint is defined as
\begin{equation}
{}^{w}n_{i,k}
=
{}^{w}j_{i,k}
\in
\mathbb{R}^{3}.
\end{equation}

Considering the residual joint torque margins, the Dynamic RFP is bounded by

\begin{equation}
\begin{aligned}
{}^{w}n_{i,k}^{T}{}^{w}f
&\leq
\alpha\tilde{\tau}_{i,k}^{+},
\\
-{}^{w}n_{i,k}^{T}{}^{w}f
&\leq
-\alpha\tilde{\tau}_{i,k}^{-},
\end{aligned}
\end{equation}
where $\tilde{\tau}_{i,k}^{+}$ and
$\tilde{\tau}_{i,k}^{-}$ denote the upper and lower
bounds of the residual joint torque, respectively,
and $0<\alpha\leq1$ scales the residual torque interval.

The proposed metric evaluates whether the entire task-force uncertainty ball is contained within the Dynamic RFP. For each joint-induced half-space constraint, the signed distance is defined as the signed distance from the ball center to the corresponding constraint plane, minus the uncertainty radius:

\begin{equation}
\label{eq:signed_distance}
\begin{aligned}
d_{i,k}^{+}
&=
\frac{
\alpha\tilde{\tau}_{i,k}^{+}
-
{}^{w}n_{i,k}^{T}{}^{w}f_k^{c}
}{
\|{}^{w}n_{i,k}\|_2
}
-r_k,
\\
d_{i,k}^{-}
&=
\frac{
-\alpha\tilde{\tau}_{i,k}^{-}
+
{}^{w}n_{i,k}^{T}{}^{w}f_k^{c}
}{
\|{}^{w}n_{i,k}\|_2
}
-r_k.
\end{aligned}
\end{equation}

Since containment of the task-force uncertainty ball requires all half-space constraints to be satisfied, the minimum signed distance is adopted as the overall force capability margin
at the $k$-th waypoint:
\begin{equation}
d_k
=
\min_{i=1,\ldots,n_a}
\left\{
\min
\left(
d_{i,k}^{+},
d_{i,k}^{-}
\right)
\right\}.
\end{equation}

The geometric interpretation of $d_k$ is
\begin{equation}
\label{eq:distance_interpretation}
\begin{aligned}
d_k>0
&\Rightarrow
\mathcal{B}_k\subset\mathcal{P}_{r,k},\\
d_k=0
&\Rightarrow
\mathcal{B}_k \text{ is tangent to } \mathcal{P}_{r,k},\\
d_k<0
&\Rightarrow
\mathcal{B}_k \text{ extends outside } \mathcal{P}_{r,k}.
\end{aligned}
\end{equation}

To obtain a numerically stable metric, a normalization scale is introduced as
\begin{equation}
\rho_k
=
\max
\left(
\left\|
{}^{w}f_k^{c}
\right\|_2
+
r_k,
\varepsilon
\right),
\end{equation}
where $\varepsilon$ is a small positive constant.

Based on the normalized violation magnitude, the force capability penalty at the $k$-th waypoint is defined as
\begin{equation}
c_k^{\mathrm{cap}}
=
\begin{cases}
0,
&
d_k \geq 0,
\\
\exp
\left(
\kappa
\frac{-d_k}{\rho_k}
\right)
-
1,
&
d_k < 0,
\end{cases}
\end{equation}
where $\kappa>0$ controls the growth rate of the penalty.

Finally, the force capability cost over the entire trajectory is
\begin{equation}
\mathcal{J}_1
=
\sum_{k=1}^{N}
c_k^{\mathrm{cap}}.
\end{equation}

Consequently, minimizing $\mathcal{J}_1$ encourages the optimizer to satisfy the nominal task-force requirement while retaining an additional safety margin specified by $r_k$. The task-force ball serves as a safety-margin construction: partial violation of its boundary does not necessarily imply
that the nominal task force is infeasible. Specifically, $d_k\geq0$ indicates that the full prescribed
safety margin is satisfied, whereas $d_k+r_k\geq0$ indicates feasibility of the nominal task force under the adopted model. The weighted objective allows the additional safety margin to be traded against other performance objectives.

\subsection{Whole-Body Trajectory Optimization}

Based on the force capability metric introduced in the previous subsection, this subsection formulates a whole-body trajectory optimization problem for the mobile manipulator. The objective is to optimize the redundant whole-body degrees of freedom while satisfying the prescribed end-effector task trajectory. In this way, the robot can maintain sufficient force capability, avoid kinematically unfavorable configurations, respect joint safety margins, and generate smooth and stable motions.

Let the whole-body configuration trajectory to be optimized be denoted by
\begin{equation}
\mathcal{Q}
=
\left\{
q_k
\right\}_{k=1}^{N},
\end{equation}
where the whole-body configuration at the $k$-th waypoint is defined as 
$q_k=\left[q_{b,k}^{T}, q_{a,k}^{T}\right]^{T}$. $q_{b,k}\in\mathbb{R}^{3}$ denotes the mobile base configuration, including the planar position and yaw angle, and $q_{a,k}\in\mathbb{R}^{n_a}$ denotes the arm joint configuration.

The overall multi-objective cost function is defined as
\begin{equation}
\mathcal{J}
=
\sum_{i=1}^{5}
w_i
\mathcal{J}_i,
\end{equation}
where $w_i\geq0$ is the weight associated with the $i$-th cost term. The five cost terms correspond to force capability, manipulability, joint-limit avoidance, trajectory smoothness, and high-frequency base oscillation suppression, respectively.

The force capability cost $\mathcal{J}_1$ is defined in the previous subsection. It penalizes violations of the task force uncertainty ball with respect to the Dynamic Residual Force Polytope (Dynamic RFP). Therefore, minimizing $\mathcal{J}_1$ encourages the optimizer to find whole-body configurations that provide sufficient force generation capability for the desired task forces. Unlike formulations that maximize force capability indiscriminately, this cost penalizes insufficient force margins while allowing other objectives to shape the solution once the task force requirement is satisfied.

Arm translational manipulability is incorporated to promote favorable arm configurations during task execution. Although the optimization variables include both the mobile base and the arm, the manipulability metric is computed using only the arm joints.

Let ${}^{w}J_{t,k}\in\mathbb{R}^{3\times n_a}$ denote the arm translational Jacobian expressed in the world frame, as defined in the force capability formulation, where $n_a=7$ for the robot considered in this work. This Jacobian maps arm joint velocities to end-effector linear velocity with the mobile base held fixed.

The regularized arm translational manipulability measure at the $k$-th waypoint is defined as
\begin{equation}
\label{eq:arm_manipulability}
\mu_k
=
\sqrt{
\det\left(
{}^{w}J_{t,k}({}^{w}J_{t,k})^{T}
+\lambda I_3
\right)
},
\end{equation}
where $\lambda>0$ is a small regularization coefficient and $I_3$ is the three-dimensional identity matrix.

The mobile base contributes indirectly to this objective by enabling different arm configurations that satisfy the prescribed end-effector pose. Thus, the framework optimizes the whole-body trajectory while evaluating the translational manipulability of the arm. The manipulability cost over the trajectory is defined as
\begin{equation}
\mathcal{J}_2
=
\sum_{k=1}^{N}
\frac{1}{\max(\mu_k,\varepsilon_m)},
\end{equation}
where $\varepsilon_m>0$ prevents division by very small values.

To prevent the arm joints from approaching their motion limits, a joint-limit avoidance cost is introduced. 
For the $i$-th arm joint, the joint center and safe half-width are defined as 
$q_{a,i}^{\mathrm{mid}}=(q_{a,i}^{\max}+q_{a,i}^{\min})/2$ and 
$h_i=\max\left(\varepsilon_h,(q_{a,i}^{\max}-q_{a,i}^{\min})/2-\delta_i\right)$, 
where $\delta_i$ is the joint safety margin and $\varepsilon_h$ is a small positive constant.

At the $k$-th waypoint, the joint-limit avoidance penalty is defined as
\begin{equation}
c_k^{\mathrm{lim}}
=
\sum_{i=1}^{n}
\left(
e_{i,k}^{2}
+
v_{i,k}^{4}
\right),
\end{equation}
where
$e_{i,k}=(q_{a,i,k}-q_{a,i}^{\mathrm{mid}})/h_i$
denotes the normalized deviation from the joint center, and
$v_{i,k}=\max\left(0,\left(|q_{a,i,k}-q_{a,i}^{\mathrm{mid}}|-h_i\right)/\delta_i\right)$
denotes the normalized violation of the predefined safe joint range. The quadratic term penalizes deviations from the joint center, while the fourth-order term imposes a stronger penalty when the joint approaches or exceeds the safe region.

The joint-limit cost over the entire trajectory is
\begin{equation}
\mathcal{J}_3
=
\sum_{k=1}^{N}
c_k^{\mathrm{lim}}.
\end{equation}

To improve temporal continuity and execution feasibility, a trajectory smoothness cost is imposed in the whole-body configuration space. 
Since the base and arm coordinates have different physical units and scales, a diagonal weighting matrix $W_s$ is introduced. 
The single-step smoothness penalty is defined as 
$\sigma_k=\frac{1}{2}\left\|W_s(q_k-q_{k-1})\right\|_2^2$, 
and the smoothness cost over the entire trajectory is given by
\begin{equation}
\mathcal{J}_4
=
\sum_{k=2}^{N}
\sigma_k.
\end{equation}
This term suppresses abrupt changes between adjacent waypoints and improves the executability of the optimized trajectory.

During mobile manipulation, the mobile base may exhibit high-frequency back-and-forth oscillations, especially when multiple redundant configurations satisfy the same end-effector pose constraint. To reduce this effect, a spectral regularization term is introduced for the base trajectory. Let
\begin{equation}
b_{\xi}
=
\left\{
b_{\xi,k}
\right\}_{k=1}^{N},
\qquad
\xi
\in
\left\{
x_b,
y_b,
\theta_b
\right\},
\end{equation}
denote the discrete base trajectory along direction $\xi$. Let $\hat{b}_{\xi,\ell}$ denote the spectral coefficient of $b_{\xi}$ at frequency index $\ell$, obtained by applying a discrete cosine transform (DCT) to the base trajectory. The high-frequency base oscillation cost is defined as
\begin{equation}
\mathcal{J}_5
=
\frac{1}{2}
\sum_{\xi\in\{x_b,y_b,\theta_b\}}
\eta_{\xi}
\sum_{\ell\in\mathcal{H}}
\gamma_{\ell}
\hat{b}_{\xi,\ell}^{2},
\end{equation}
where $\mathcal{H}$ denotes the set of frequency indices higher than a predefined cutoff frequency, $\gamma_{\ell}$ is the frequency-dependent weighting coefficient, and $\eta_{\xi}$ is the weight for each base motion direction. This term penalizes high-frequency components in the base trajectory and improves the stability of the resulting whole-body motion.

The complete whole-body trajectory optimization problem is formulated as
\begin{equation}
\begin{aligned}
\mathcal{Q}^{*}
=
\arg\min_{\mathcal{Q}}
\quad
&
\mathcal{J}(\mathcal{Q})
\\
\mathrm{s.t.}
\quad
&
x(q_k)
=
x_k,
\qquad
\qquad
k=1,\ldots,N,
\\
&
q_{a}^{\min}
\leq
q_{a,k}
\leq
q_{a}^{\max},
\qquad
k=1,\ldots,N,
\\
&
q_1
=
q_{\mathrm{init}},
\end{aligned}
\end{equation}
where $x(q_k)$ denotes the forward kinematics of the end-effector pose under the whole-body configuration $q_k$, and $x_k$ is the desired end-effector pose specified by the task trajectory. The joint-limit constraints ensure that the optimized arm trajectory remains within the feasible joint range, while the initial condition anchors the optimized trajectory to the measured initial whole-body configuration.

The weights $w_i$ are kept fixed across payload conditions. Task adaptation is achieved through the force capability cost $\mathcal{J}_1$, which responds to changes in the desired task force. For tasks with lower force demands, once the task force uncertainty ball is already contained within the Dynamic RFP, the optimizer can place more emphasis on manipulability, joint safety, trajectory smoothness, and base stability. As a result, the proposed formulation does not simply maximize force capability, but instead seeks a task-sufficient force margin while preserving favorable whole-body motion properties.

\section{Experiments}

The proposed framework is evaluated through a VLM mass-estimation benchmark, a MuJoCo simulation, and real-robot experiments. The real-robot experiments are conducted on the Mobile Collaborative Robot Assistant (MOCA), which consists of a mobile base and a 7-DoF Franka Emika robotic arm. A basket and an Intel RealSense D435i RGB-D camera are rigidly mounted on the end-effector for object transportation and visual perception. The basket, camera, and mounting attachments are included in the robot dynamics model. The VLM estimates only the beverage payload mass, whose contribution is accounted for separately
through the desired task force.

The Franka arm is controlled through a torque interface, while the mobile base is commanded through a velocity interface. To prevent actuator saturation and ensure safe operation, conservative torque limits are imposed on the arm joints:
\begin{equation}
\tau_{i,\max}
=
\begin{cases}
60.0\,\mathrm{Nm}, & i=1,\ldots,4,\\
12.0\,\mathrm{Nm}, & i=5,\ldots,7.
\end{cases}
\end{equation}
These limits are incorporated into the residual force polytope computation and trajectory optimization. The task-force uncertainty radius is fixed at $r_k=10\,\mathrm{N}$ for all waypoints to provide an additional safety margin. The objective weights of the proposed method are kept fixed across payload conditions, so its task-adaptive behavior arises from changes in the task-force requirement rather than manual weight adjustment. Throughout the experimental evaluation, manipulability refers to the arm translational manipulability defined in Eq.~\eqref{eq:arm_manipulability}.

The evaluation consists of four parts. First, the VLM-based mass estimator is independently evaluated on the Household Test Set~\cite{image2mass} to verify the accuracy of the physical-property estimation used for task-force prediction. Second, a continuous-trajectory simulation with time-varying task forces evaluates the task-adaptive trade-off between directional force capability and arm translational manipulability under continuously changing force magnitudes and directions. Third, a lifting trajectory experiment evaluates the proposed method during dynamic manipulation under light and heavy payload conditions. Finally, a single-point holding experiment compares different redundancy-resolution strategies at the same fixed end-effector pose and further analyzes their residual force polytopes and arm translational velocity manipulability ellipsoids.

\subsection{VLM-based Mass Estimation Validation}

To independently evaluate the mass estimation capability of the VLM, we conduct experiments on the Household Test Set introduced in Image2Mass~\cite{image2mass}, which contains 56 household objects and 423 RGB images. We use GPT-4o as the VLM for single-view object mass estimation and compare our estimator with both the single-view and multi-view variants of Image2Mass. The quantitative results are summarized in Table~\ref{tab:vlm_mass_estimation}, where $\downarrow$ ($\uparrow$) indicates that a lower (higher) value corresponds to better estimation performance.

As shown in Table~\ref{tab:vlm_mass_estimation}, our method achieves the best performance across all five evaluation metrics. In particular, using only a single RGB image, our method obtains an mALDE of 0.333 and an mAPE of 0.377, compared with 0.437 and 0.637, respectively, for the multi-view Image2Mass method. Moreover, our method achieves an $r^2_{\mathrm{ls}}$ of 0.839, an mMnRE of 0.742, and a $q$ score of 0.917, consistently outperforming both Image2Mass variants. These results demonstrate that the VLM can provide reliable mass estimates from single-view RGB observations, supporting its use for task-force estimation in the proposed framework.

\begin{table}[t]
\centering
\caption{Mass estimation performance on the Household Test Set
(56 items, 423 images). The Image2Mass results are taken from~\cite{image2mass}.}
\label{tab:vlm_mass_estimation}
\resizebox{\columnwidth}{!}{
\begin{tabular}{lccccc}

\noalign{\hrule height 1.2pt}
Method
& mALDE $\downarrow$
& mAPE $\downarrow$
& $r^2_{\mathrm{ls}}$ $\uparrow$
& mMnRE $\uparrow$
& $q$ $\uparrow$ \\
\hline
Image2Mass (single-view)
& 0.465
& 0.685
& 0.691
& 0.675
& 0.766 \\

Image2Mass (multi-view)
& 0.437
& 0.637
& 0.701
& 0.693
& 0.786 \\

\textbf{Ours (single-view)}
& $\mathbf{0.333}$
& $\mathbf{0.377}$
& $\mathbf{0.839}$
& $\mathbf{0.742}$
& $\mathbf{0.917}$ \\
\noalign{\hrule height 1.2pt}
\end{tabular}
}
\end{table}

\subsection{Continuous Trajectory with Time-Varying Task Force}

To evaluate the adaptability of the proposed method under continuously varying motion and force requirements, a MuJoCo simulation is conducted over $T=12~\mathrm{s}$. The end-effector orientation is fixed, while both the desired position and task force vary continuously according to
\begin{equation}
\mathbf{y}(t)
=
w(t)
\left(
\mathbf{c}
+
\mathbf{a}\odot
\sin(2\pi \mathbf{f}t+\boldsymbol{\phi})
\right),
\end{equation}
where $\odot$ denotes element-wise multiplication. For the position trajectory, $\mathbf{y}(t)=\mathbf{p}_d(t)-\mathbf{p}_0$ and $\mathbf{c}=\mathbf{0}$; for the task force, $\mathbf{y}(t)=\mathbf{F}_d(t)$ and $\mathbf{c}=\mathbf{b}$. The smooth window function is defined as $w(t)=4s(t)(1-s(t))$, where $s(t)=10\tau^3-15\tau^4+6\tau^5$ and $\tau=t/T$. The corresponding parameters are listed in Table~\ref{tab:continuous_sim_parameters}.

\begin{table}[t]
\centering
\caption{Parameters of the continuous trajectory and task-force simulation.}
\label{tab:continuous_sim_parameters}
\resizebox{\columnwidth}{!}{
\begin{tabular}{lll}
\noalign{\hrule height 1.2pt}
Parameter & Position trajectory & Task force \\
\hline
Offset $\mathbf{c}$
& $\mathbf{0}$
& $[25,\ 15,\ -65]^\mathsf{T}\,\mathrm{N}$ \\

Amplitude $\mathbf{a}$
& $[0.10,\ 0.10,\ -0.10]^\mathsf{T}\,\mathrm{m}$
& $[80,\ 60,\ 80]^\mathsf{T}\,\mathrm{N}$ \\

Frequency $\mathbf{f}$
& $[0,\ 0.20,\ 0.10]^\mathsf{T}\,\mathrm{Hz}$
& $[1/6,\ 1/4,\ 1/12]^\mathsf{T}\,\mathrm{Hz}$ \\

Phase $\boldsymbol{\phi}$
& $[\pi/2,\ 0,\ \pi/2]^\mathsf{T}$
& $[0.5,\ \pi/2,\ \pi/4]^\mathsf{T}$ \\

Duration
& \multicolumn{2}{c}{$T=12~\mathrm{s}$} \\

Trials
& \multicolumn{2}{c}{6 per method} \\
\noalign{\hrule height 1.2pt}
\end{tabular}
}
\end{table}

Four methods are compared: Ours, Manipulability only, RFP inscribed radius, and RFP Cone. All methods use the same robot model, initial configuration, trajectory, task-force profile, and constraints. Each method is run six times under identical experimental conditions, with negligible variation across runs. The evaluated quantities are the RFP force capability along the instantaneous task-force direction and the corresponding manipulability.

\begin{figure}[!htbp]
    \centering
    \includegraphics[width=1.0\linewidth]{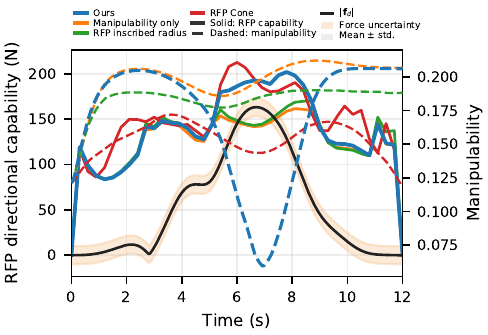}
    \caption{Directional force capability and arm translational manipulability under continuously varying task-force demands. Solid and dashed curves represent the RFP directional capability and manipulability, respectively. The black curve denotes the magnitude of the desired task force, and the shaded region represents the force uncertainty.}
    \label{fig:sim_dir}
\end{figure}

As shown in Fig.~\ref{fig:sim_dir}, the desired force demand increases significantly during approximately $5$--$8~\mathrm{s}$. In this interval, the proposed method increases the force capability along the instantaneous task-force direction, while allowing manipulability to decrease when necessary. When the force demand becomes lower, the manipulability of Ours recovers and remains close to that of the Manipulability-only method. This behavior demonstrates that the proposed method adaptively balances force capability and kinematic manipulability according to the instantaneous task requirement.

\subsection{Lifting Trajectory Experiment}

To validate the effectiveness of the proposed method under lifting tasks with different payloads, a lifting trajectory experiment was conducted.
\begin{figure}[!htbp]
    \centering
    \includegraphics[width=0.98\linewidth]{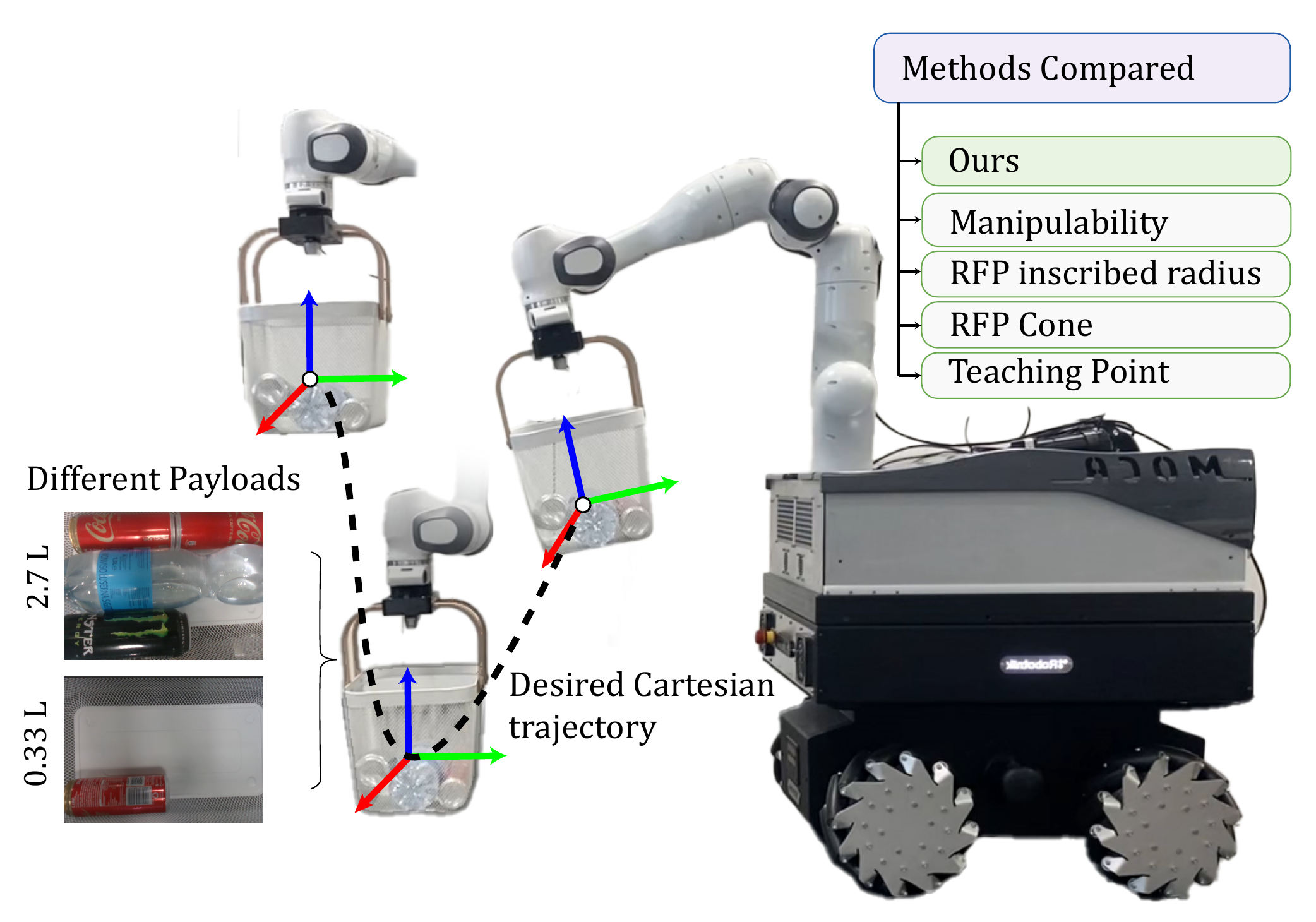}
    \caption{
    Overview of the experimental setup for the lifting and single-point holding experiments, including the mobile manipulator, payload conditions, and compared methods. The end-effector follows the indicated Cartesian trajectory in the lifting experiment and maintains a fixed desired Cartesian pose in the single-point holding experiment.
    }
    \label{fig:lifting_experiment_overview}
\end{figure}

\begin{figure*}[!htbp]
    \centering

    \includegraphics[width=0.9\textwidth]{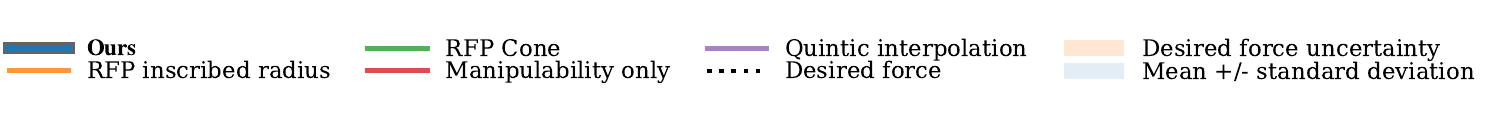}

    \setlength{\tabcolsep}{2pt}
    \renewcommand{\arraystretch}{1.0}

    \begin{tabular}{c c c}
        \raisebox{0.0cm}{\rotatebox{90}{\scriptsize RFP directional capability (N)}}
        &
        \includegraphics[width=0.46\textwidth]{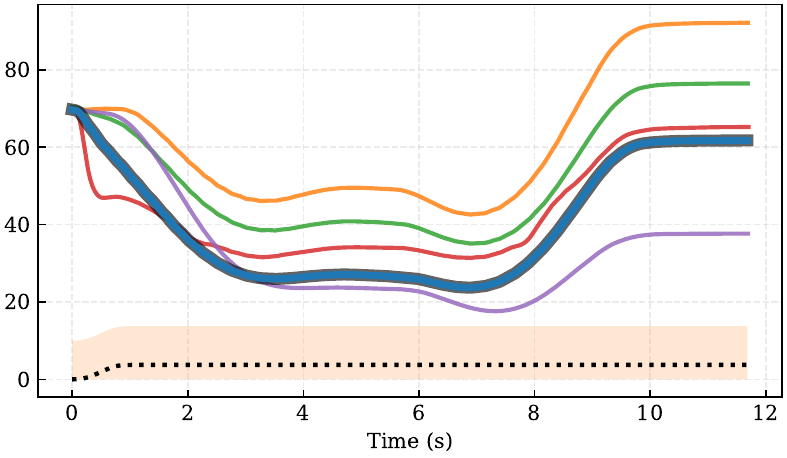}
        &
        \includegraphics[width=0.46\textwidth]{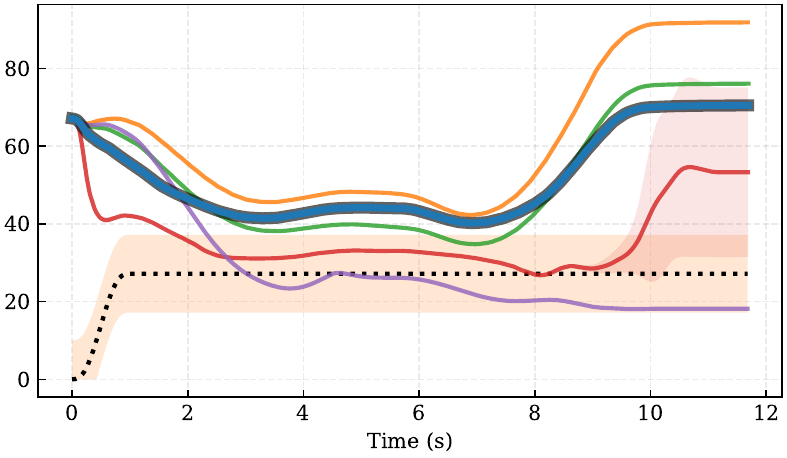}

        \\
        \raisebox{1.0cm}{\rotatebox{90}{\scriptsize manipulability}}
        &
        \includegraphics[width=0.46\textwidth]{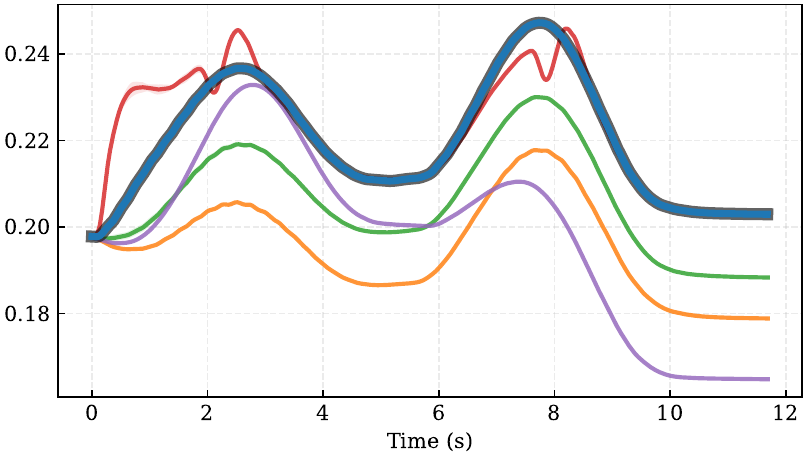}
        &
        \includegraphics[width=0.46\textwidth]{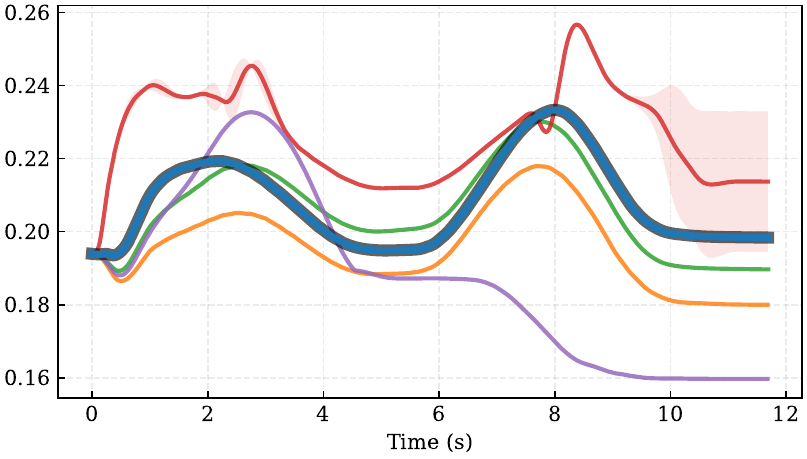}

        \\[-2pt]

        &
        \scriptsize 0.33 L beverage
        &
        \scriptsize 2.7 L beverage

    \end{tabular}

    \caption{Performance comparison of different planning methods in the 2.7~L and 0.33~L beverage lifting task.
The upper plot shows the directional dynamic residual force polytope capacity, defined as the force magnitude at the intersection between the ray along the desired force direction and the boundary of the dynamic residual force polytope.
The lower plot shows the arm manipulability.
All curves report the mean over three trials, the shaded regions denote the standard deviation, and the orange band represents the uncertainty range of the desired task force.}
    \label{fig:lifting_experiment}
\end{figure*}

In this experiment, the robot end-effector starts from the home pose and sequentially passes through two predefined waypoints, thereby lowering and lifting the object placed in the basket.
Both payload conditions were tested under the same desired Cartesian trajectory, while different planning methods were used to generate the corresponding whole-body trajectories.
Fig.~\ref{fig:lifting_experiment_overview} illustrates the experimental setup, including the desired Cartesian trajectory, the tested payload conditions, and the compared planning methods.

We analyze the maximum force capability along the vertical upward task-force direction and the arm manipulability under different payload conditions.
Here, the maximum force capability is defined as the force magnitude corresponding to the intersection between the ray along the desired force direction and the boundary of the dynamic residual force polytope.
This metric quantifies the remaining force capability of the robot along the task-force direction.
The manipulability is evaluated using Yoshikawa's manipulability measure~\cite{Yoshikawa1985Manipulability}, which describes the kinematic flexibility of the manipulator.
A larger manipulability value indicates higher local motion flexibility~\cite{Dietrich2015}.

As shown in the right column of Fig.~\ref{fig:lifting_experiment}, we first compare different planning methods when lifting bottled drinks with a total volume of 2.7 L.
The upper plot shows the directional dynamic residual force polytope capacity along the vertical upward task-force direction, while the lower plot shows the arm manipulability.
All curves represent the mean over three trials, the shaded regions denote the standard deviation, and the orange band represents the uncertainty range of the desired task force.
The directional force capacity of the proposed method remains above the upper bound of the desired force uncertainty band throughout the evaluated trajectory, indicating sufficient capacity for the nominal task force with an additional margin along the task-force direction.

In contrast, the manipulability-only optimization method improves the kinematic flexibility of the manipulator but fails to consistently maintain sufficient force capability in this high-load lifting task.
Around $8\,\mathrm{s}$, its vertical force capability even drops below the required task force.
The increased variance after this period is caused by one of the three trials failing to lift the object at approximately $8\,\mathrm{s}$, which leads to larger fluctuations in the experimental results.
This demonstrates that optimizing manipulability alone is insufficient for high-force lifting tasks.
In addition, the method using only quintic polynomial interpolation without optimization fails to lift the object at approximately $2.5\,\mathrm{s}$ and remains below the required force afterwards, resulting in task failure.

\begin{figure*}[!htbp]
    \centering

    \includegraphics[width=0.75\textwidth]{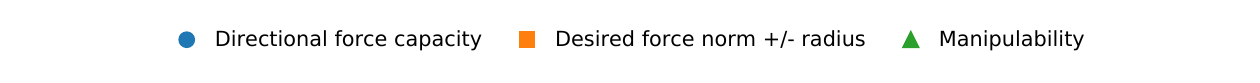}
    \vspace{2pt}

    \begin{subfigure}[b]{0.48\linewidth}
        \centering
        \includegraphics[width=\linewidth]{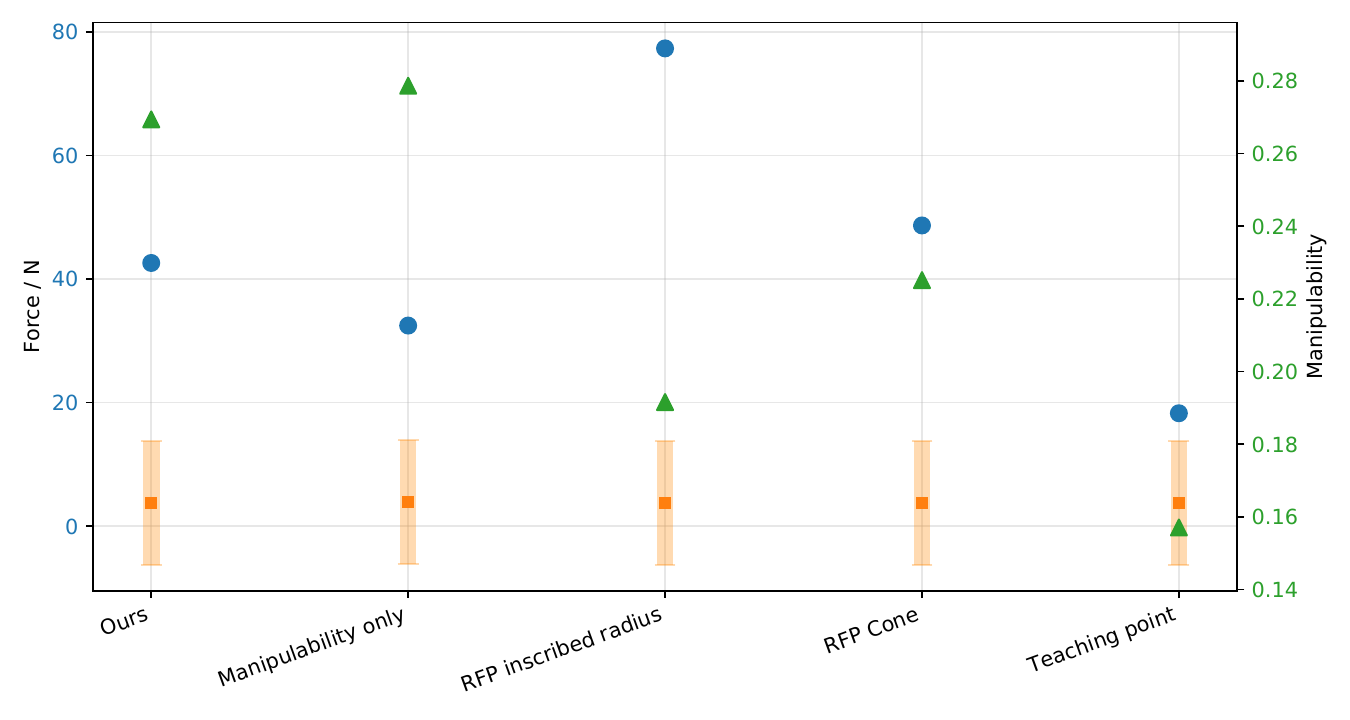}
        \caption{0.33 L beverage}
        \label{fig:single_point_033l_metrics}
    \end{subfigure}
    \hfill
    \begin{subfigure}[b]{0.48\linewidth}
        \centering
        \includegraphics[width=\linewidth]{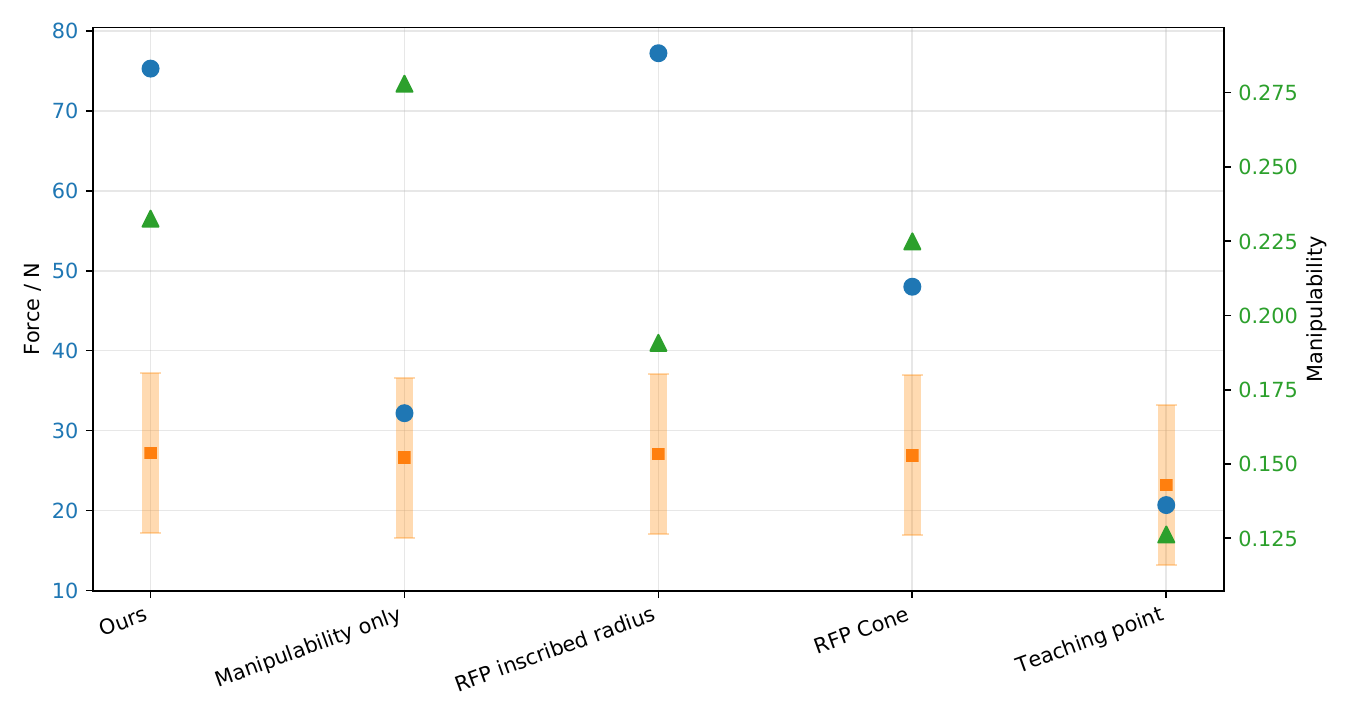}
        \caption{2.7 L beverage}
        \label{fig:single_point_2p7l_metrics}
    \end{subfigure}

    \caption{
    Single-point comparison of force output capability and manipulability under different payload conditions.
    The left vertical axis represents the directional residual force polytope capacity and the desired force range, while the right vertical axis represents the manipulability.
    }
    \label{fig:onepoint}
\end{figure*}

We further compare the proposed method with two residual-force-polytope-based methods proposed by Ferrolho et al., namely RFP inscribed radius and RFP Cone~\cite{ferrolhoResidualForcePolytope2021a}.
Both methods directly optimize force-related metrics and are therefore able to complete the 2.7 L lifting task with sufficient force capability.
It should be noted that the absolute magnitude of the force capability curves alone should not be interpreted as a direct indication of method superiority, since changing the optimization weights can shift the overall level of the curves produced by the RFP inscribed radius and RFP Cone methods~\cite{ferrolhoResidualForcePolytope2021a}.
The key difference lies in whether the method can balance force capability and motion flexibility according to the actual task requirement.

As shown in the left column of Fig.~\ref{fig:lifting_experiment}, when the payload is reduced to a single 330 mL Coke bottle, the required vertical lifting force becomes significantly smaller.
In this low-load case, the two RFP-based methods proposed by Ferrolho et al. still tend to maintain large force capability in the specified direction because their objectives mainly optimize residual-force-polytope-related quantities.
However, such a high force margin is unnecessary for a low-load lifting task and may consume the robot's configuration redundancy, preventing it from fully exploiting its kinematic flexibility.
In contrast, the proposed method adapts its optimization behavior according to the desired task force.
When the required force is small, it only needs to keep the force capability above the task requirement and its uncertainty range, while allocating more optimization freedom to improving the arm manipulability.
Consequently, in the 330 mL Coke lifting task, the manipulability achieved by the proposed method is comparable to that obtained by the manipulability-only optimization method, while still satisfying the force requirement.
These results show that the proposed method can provide sufficient force capability in high-load tasks and avoid excessive force-margin optimization in low-load tasks, thereby achieving a more task-adaptive balance between force capability and motion flexibility.

\subsection{Single-Point Holding Experiment}

\begin{figure*}[h]

    \centering

    \setlength{\tabcolsep}{2pt}
    \renewcommand{\arraystretch}{1.0}

    \begin{tabular}{c c c c c c}
\rotatebox{90}{\scriptsize 2.7 L beverage}
        &
        \includegraphics[width=0.18\textwidth]{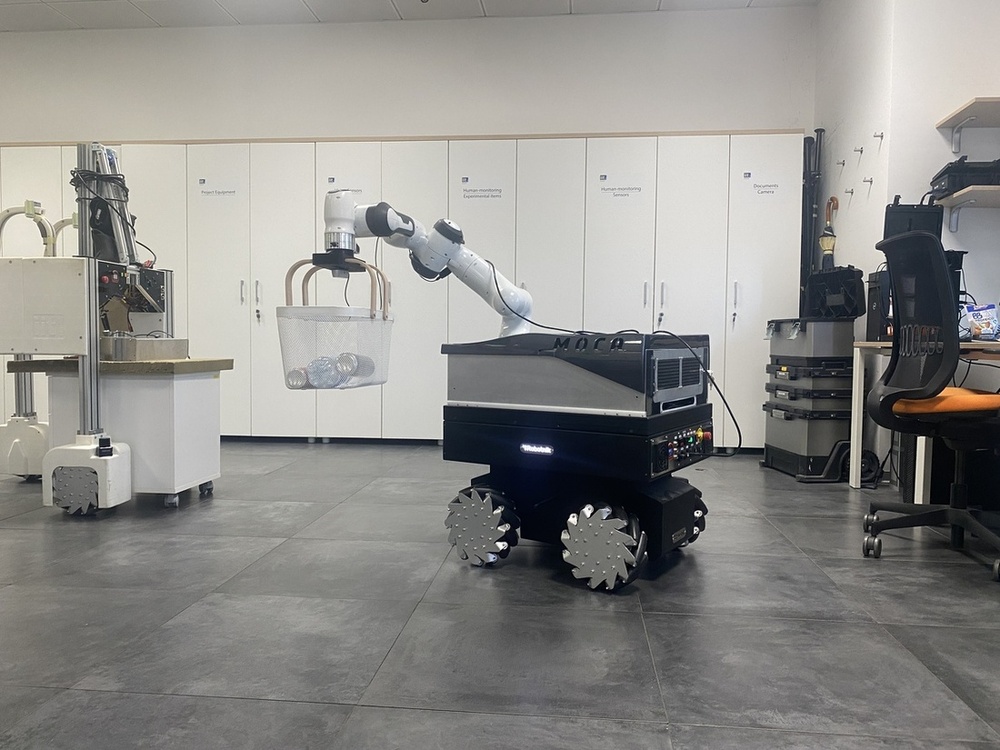}
        &
        \includegraphics[width=0.18\textwidth]{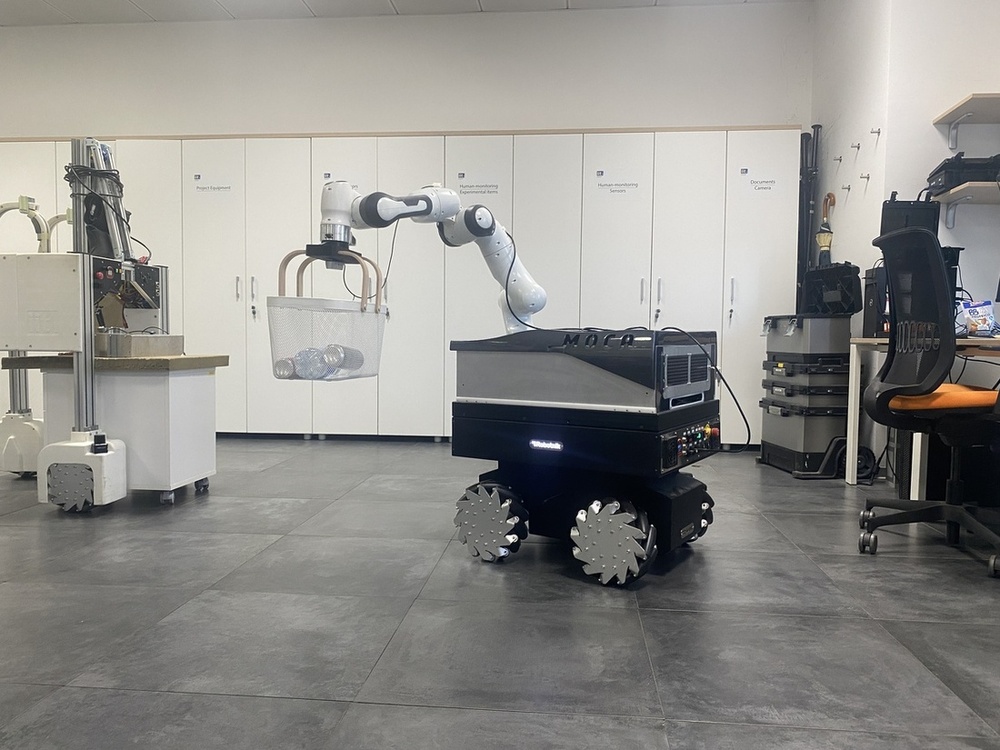}
        &        

        \includegraphics[width=0.18\textwidth]{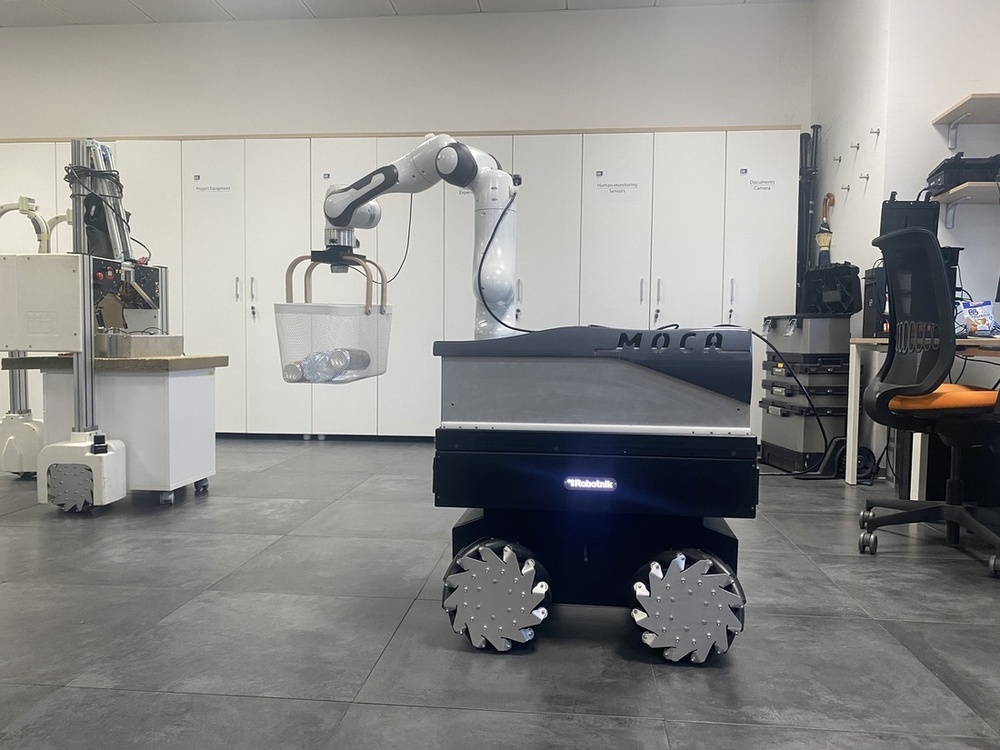}
                &
        \includegraphics[width=0.18\textwidth]{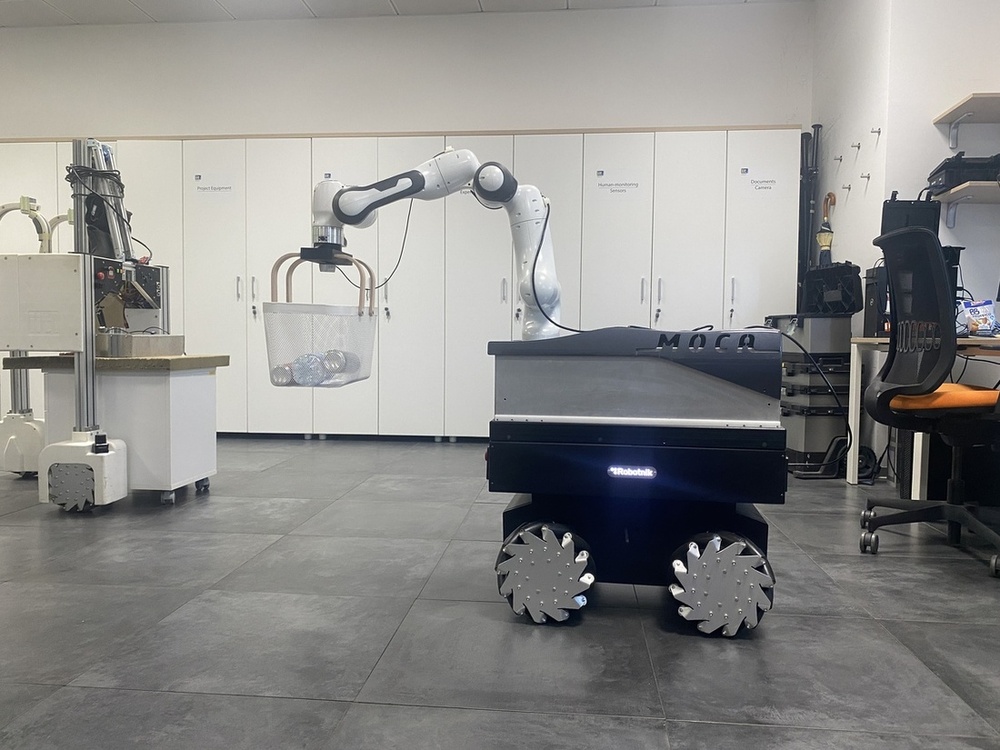}
                &
        \includegraphics[width=0.18\textwidth]{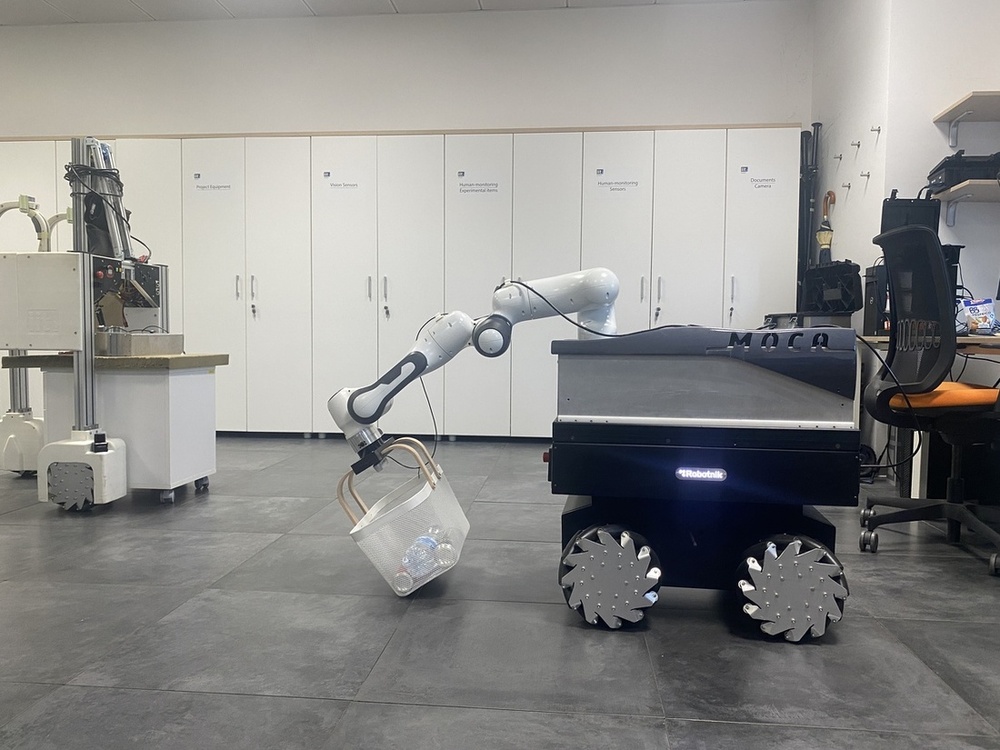}

\\
\rotatebox{90}{\scriptsize 0.33 L beverage}

        &
        \includegraphics[width=0.18\textwidth]{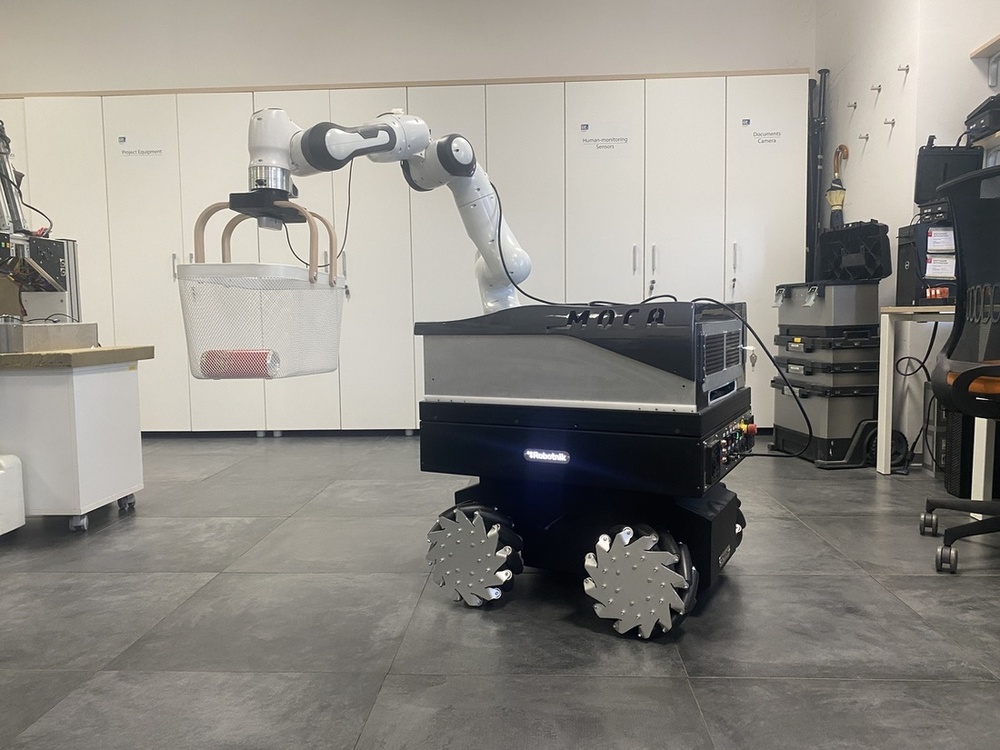}
        &
        \includegraphics[width=0.18\textwidth]{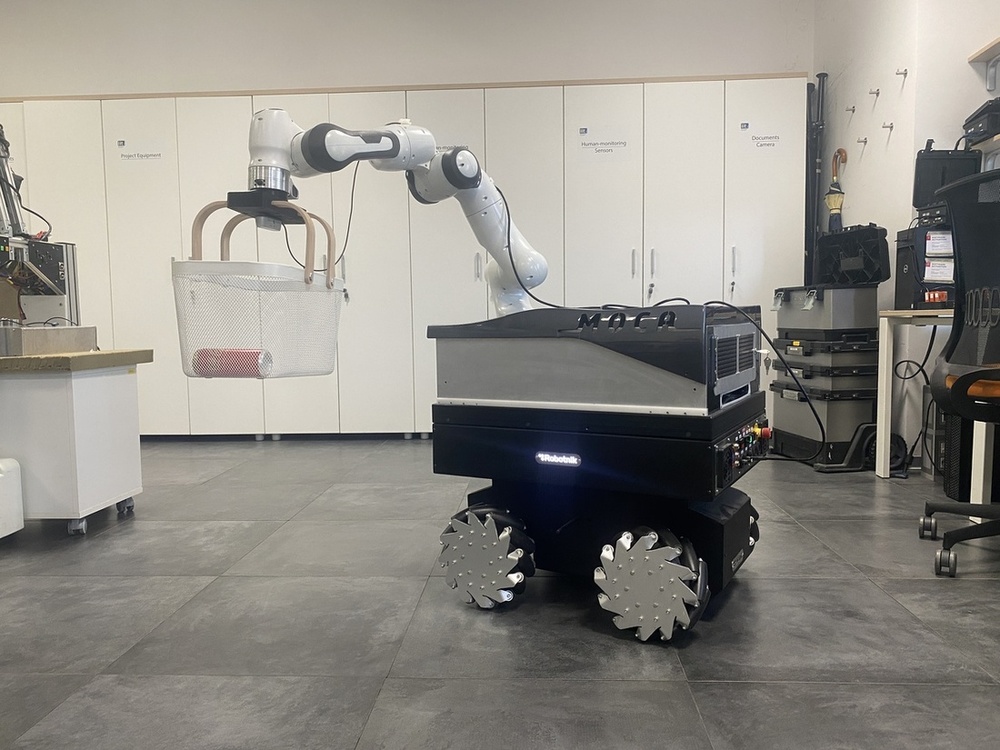}
        &        

        \includegraphics[width=0.18\textwidth]{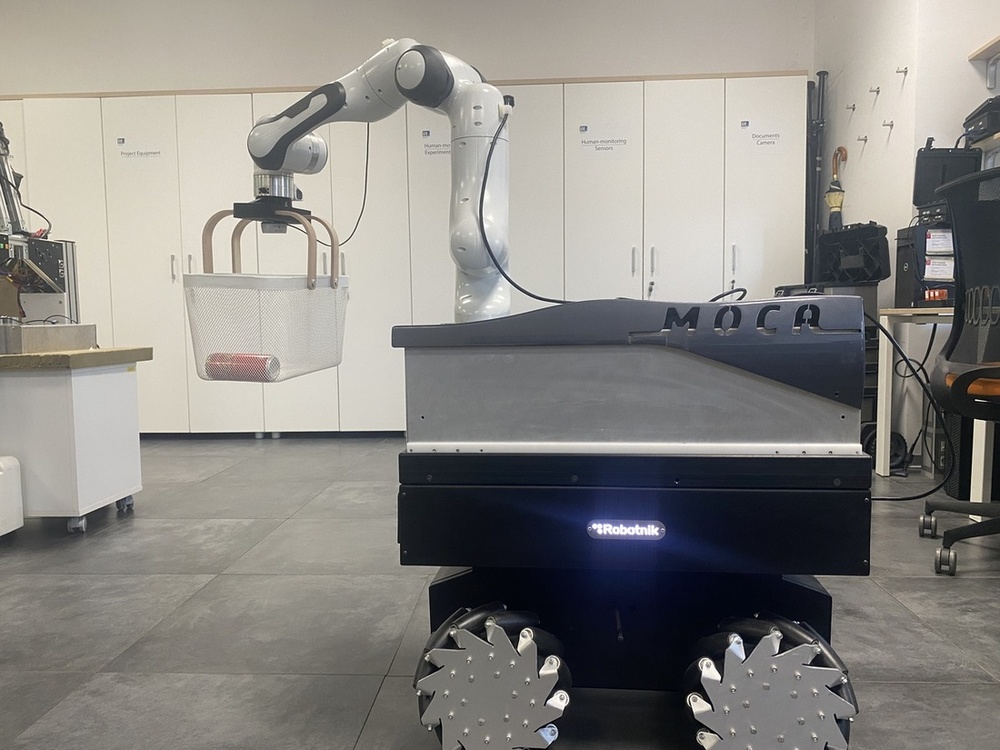}
                &
        \includegraphics[width=0.18\textwidth]{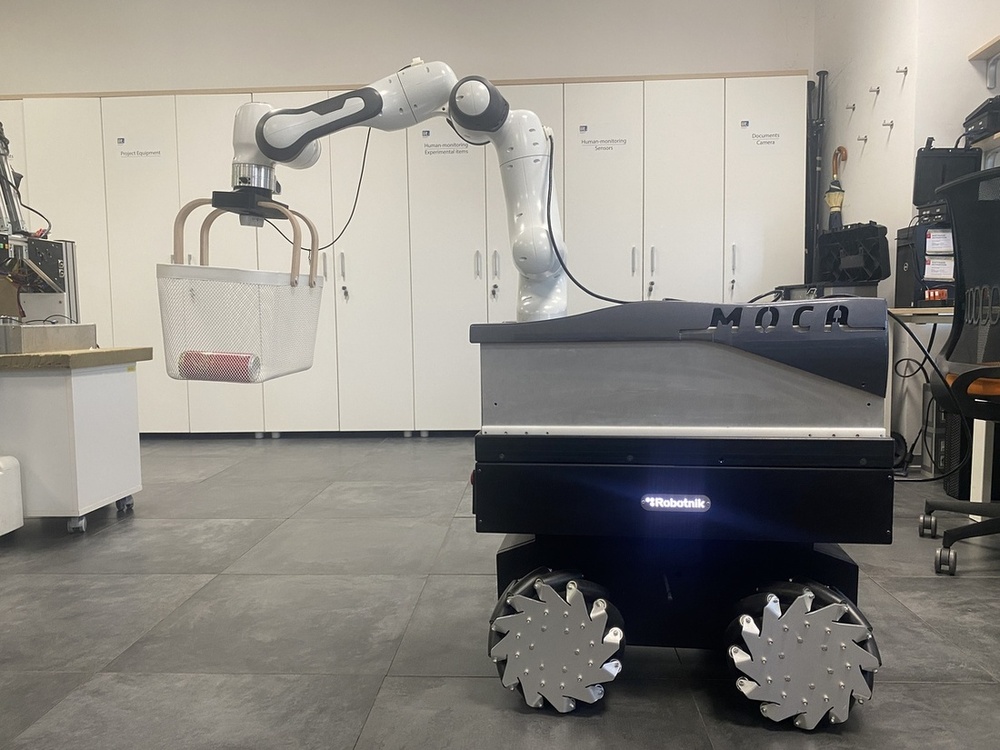}
                &
        \includegraphics[width=0.18\textwidth]{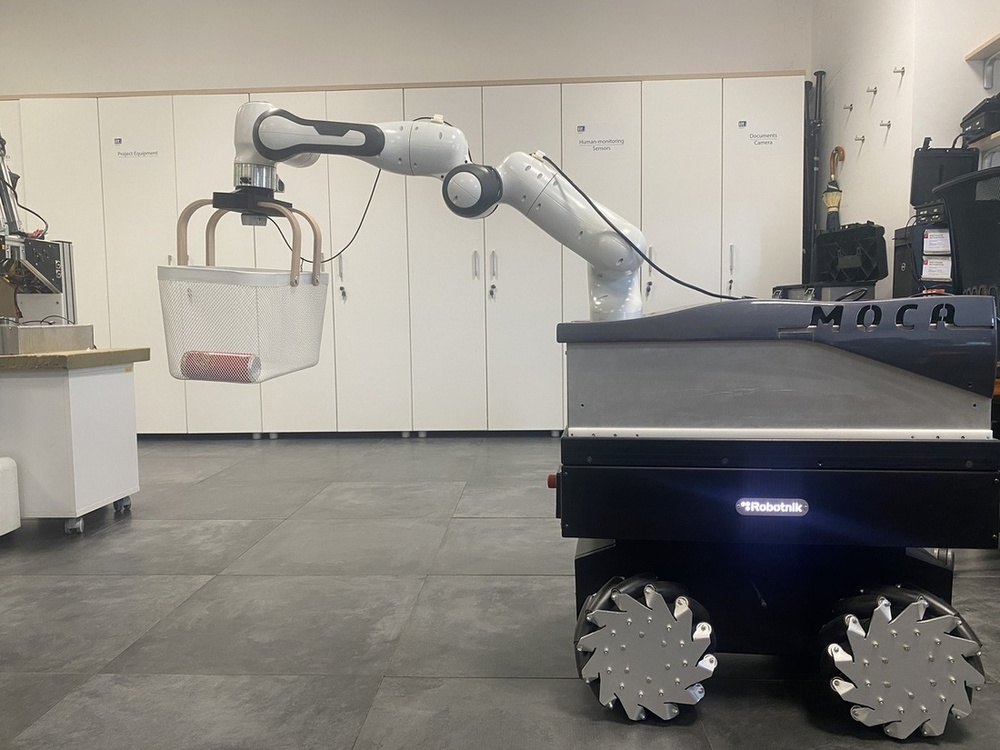}

        \\

        &
        \scriptsize Ours
        &
        \scriptsize Manipulability
        &
        \scriptsize RFP inscribed radius
        &
        \scriptsize RFP Cone
        &
        \scriptsize Teaching point
    \end{tabular}

    \caption{ 
    Comparison of real-world single-point optimization results obtained by different optimization methods under 0.33 L and 2.7 L payload conditions. All methods optimize the same end-effector pose, while producing different robot configurations due to their respective optimization objectives.
    }
    \label{fig:single_point_realrobot}
\end{figure*}

\begin{figure*}[h]

    \centering

    \includegraphics[width=0.75\textwidth]{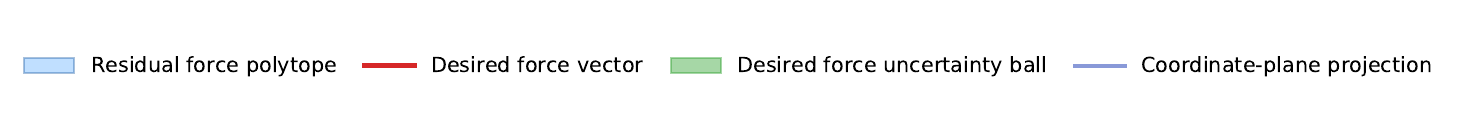}
    \vspace{2pt}

    \setlength{\tabcolsep}{2pt}
    \renewcommand{\arraystretch}{1.0}

    \begin{tabular}{c c c c c c}
        \rotatebox{90}{\scriptsize 2.7 L beverage}
        &
        \includegraphics[width=0.18\textwidth]{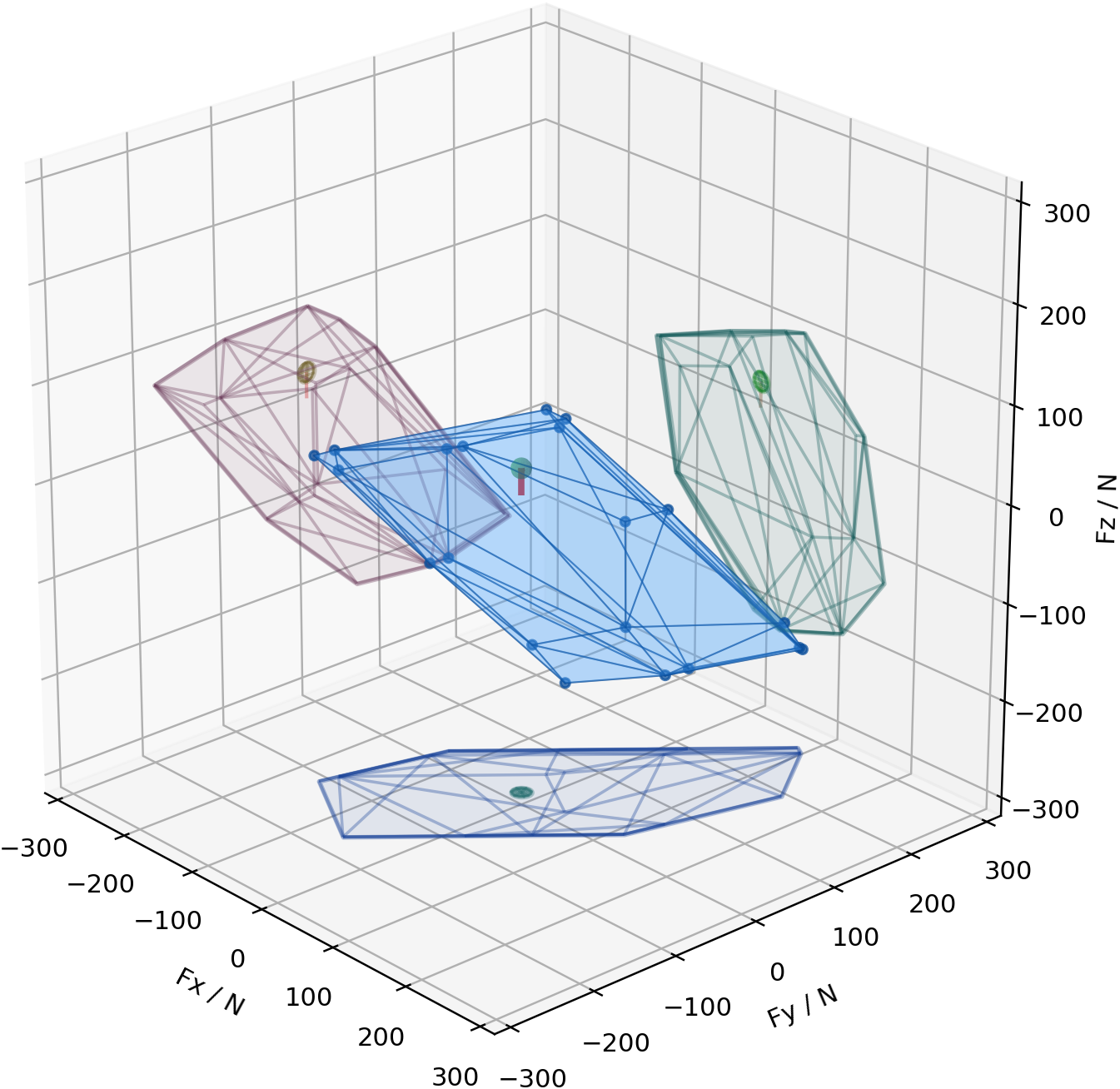}
        &
        \includegraphics[width=0.18\textwidth]{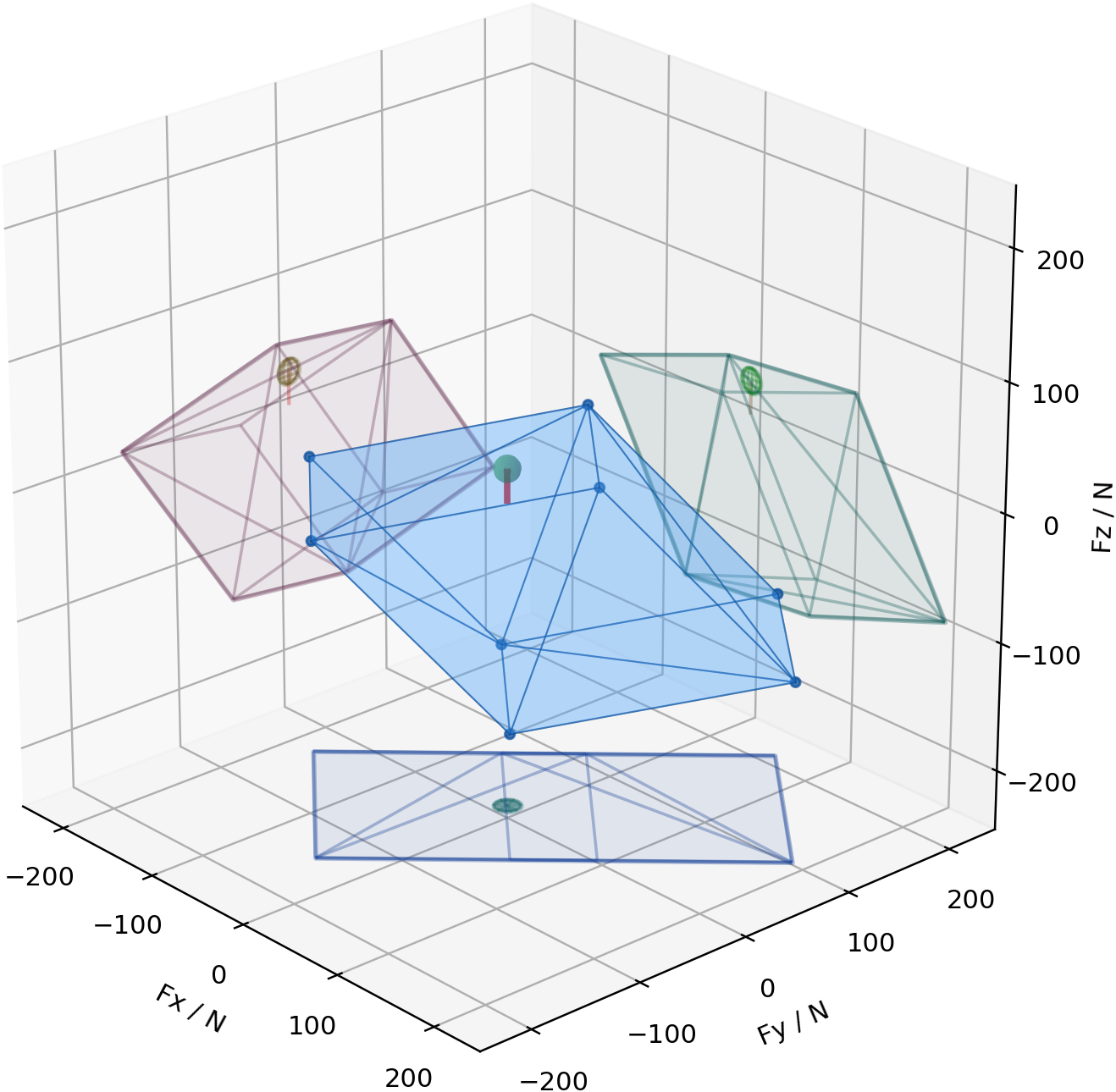}
        &        
        \includegraphics[width=0.18\textwidth]{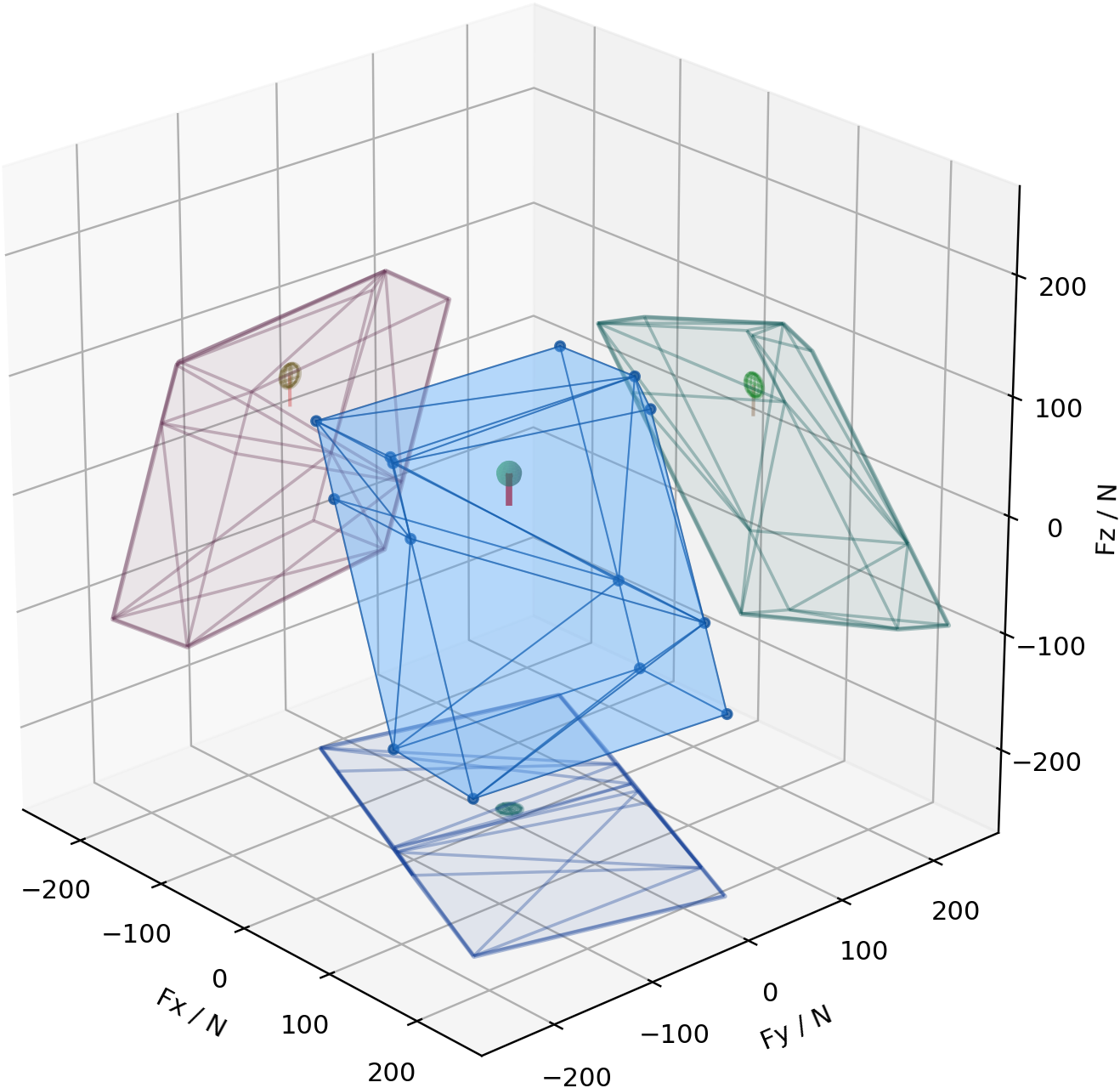}
        &        
        \includegraphics[width=0.18\textwidth]{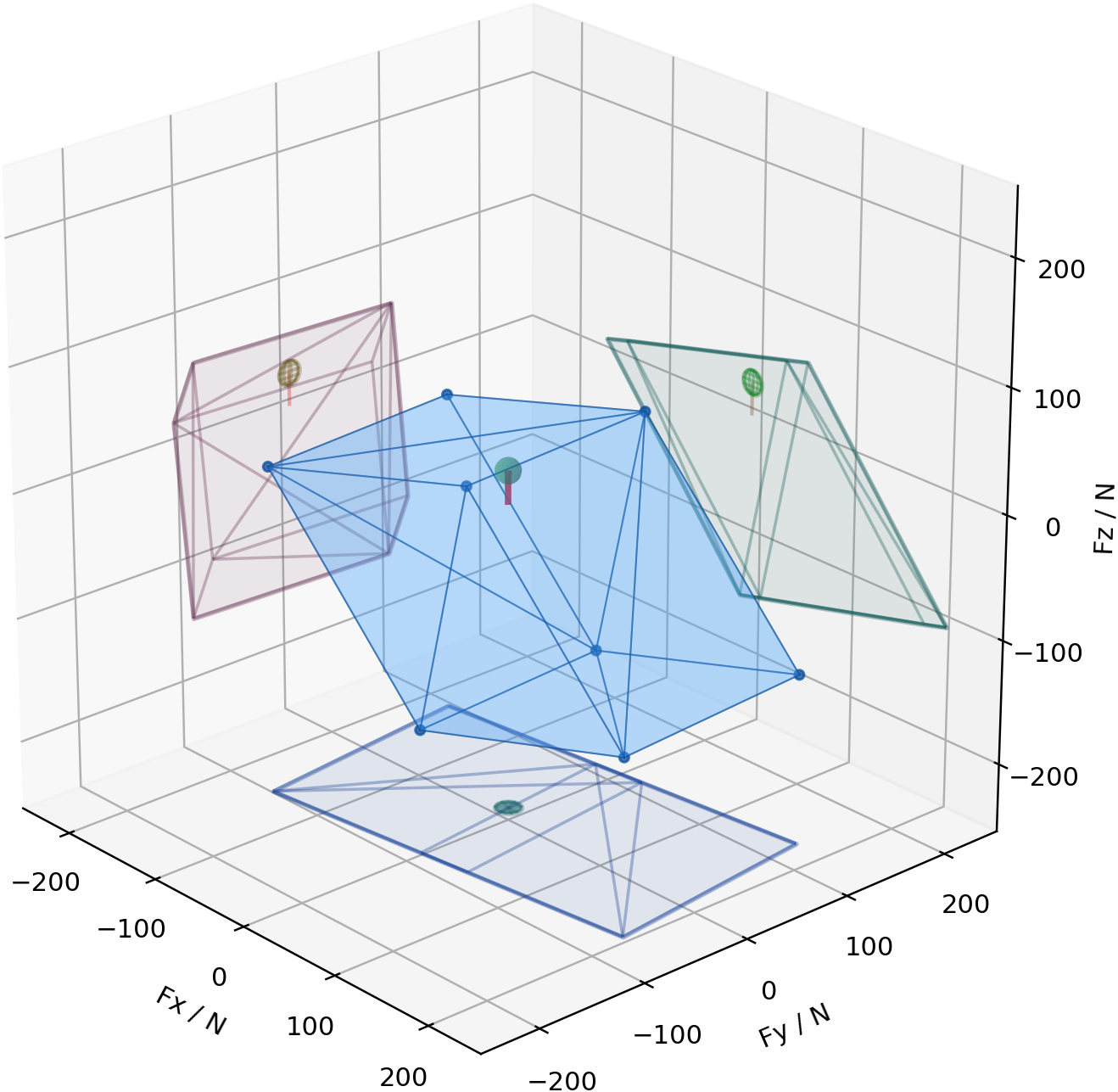}
        &        
        \includegraphics[width=0.18\textwidth]{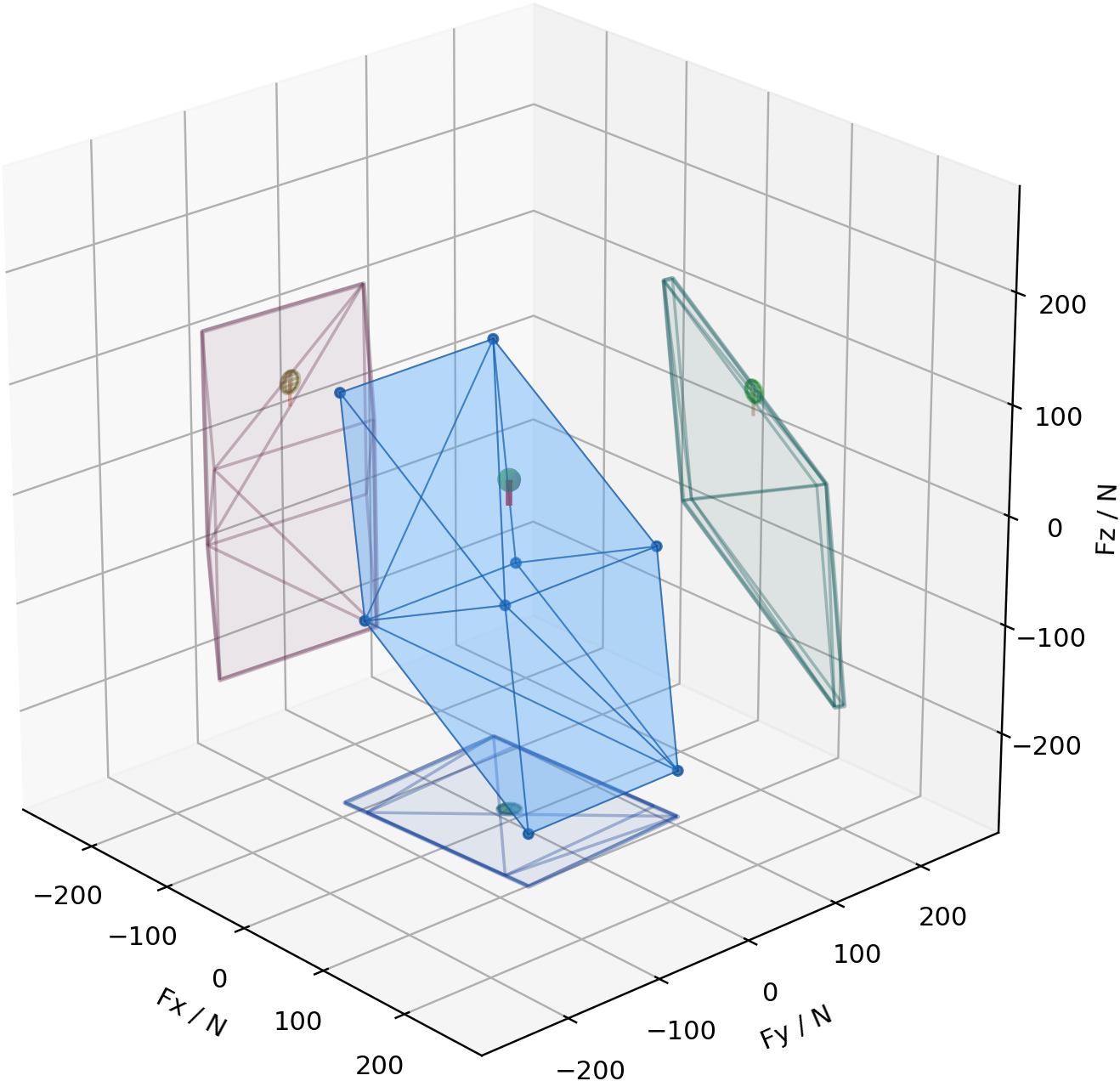}

        \\[-2pt]

        \rotatebox{90}{\scriptsize 0.33 L beverage}
        &
        \includegraphics[width=0.18\textwidth]{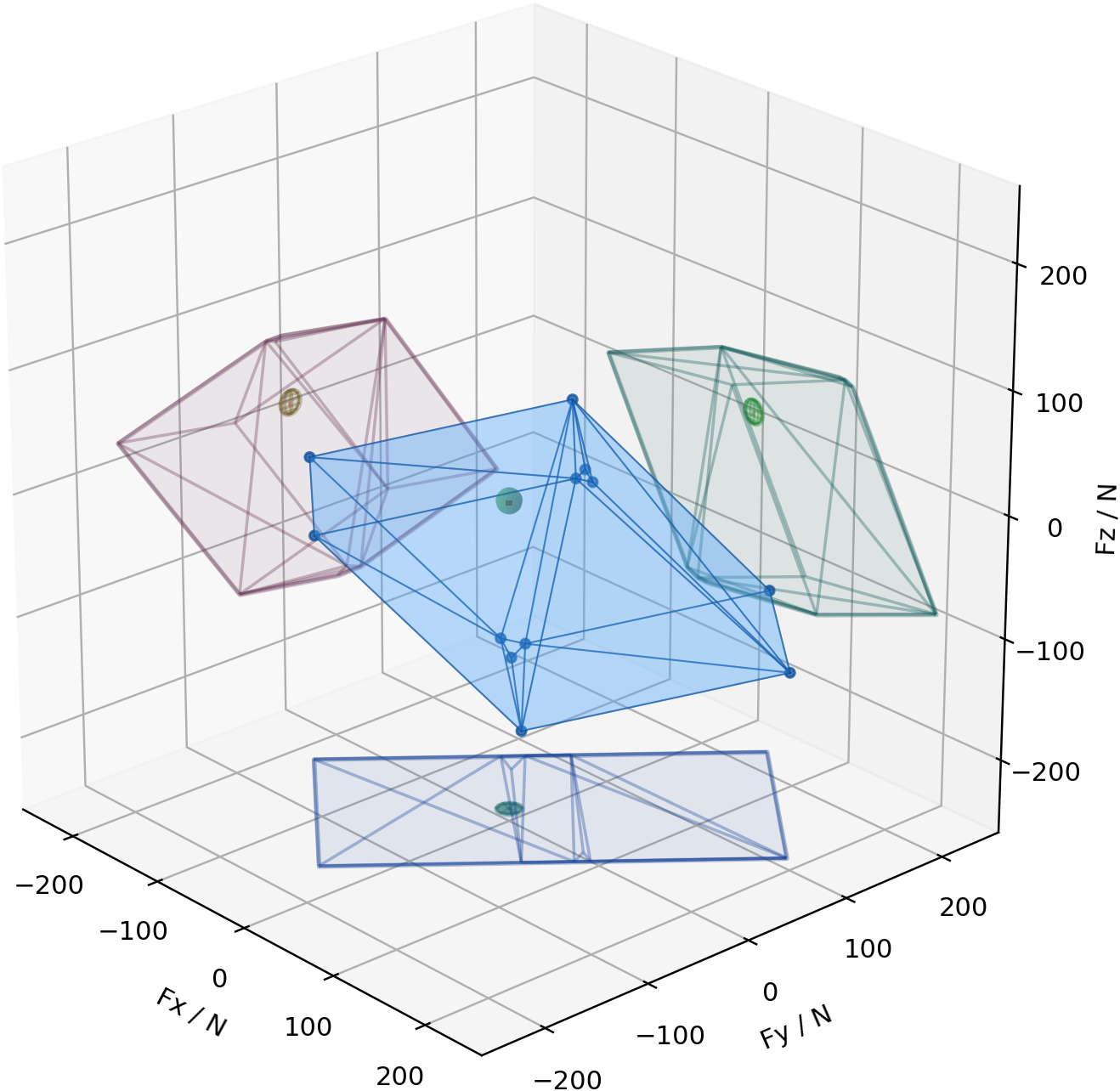}
        &
        \includegraphics[width=0.18\textwidth]{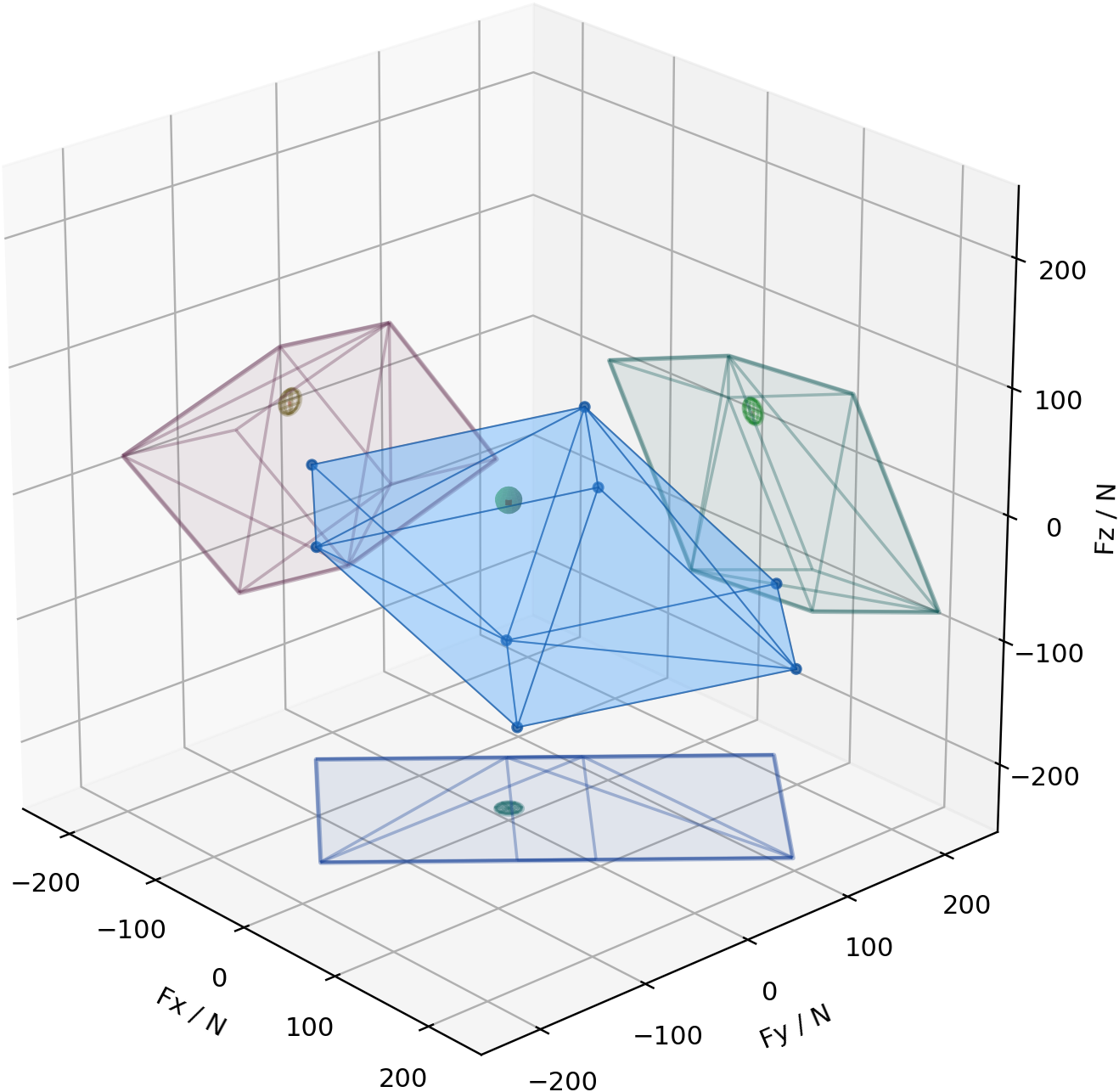}
        &
        \includegraphics[width=0.18\textwidth]{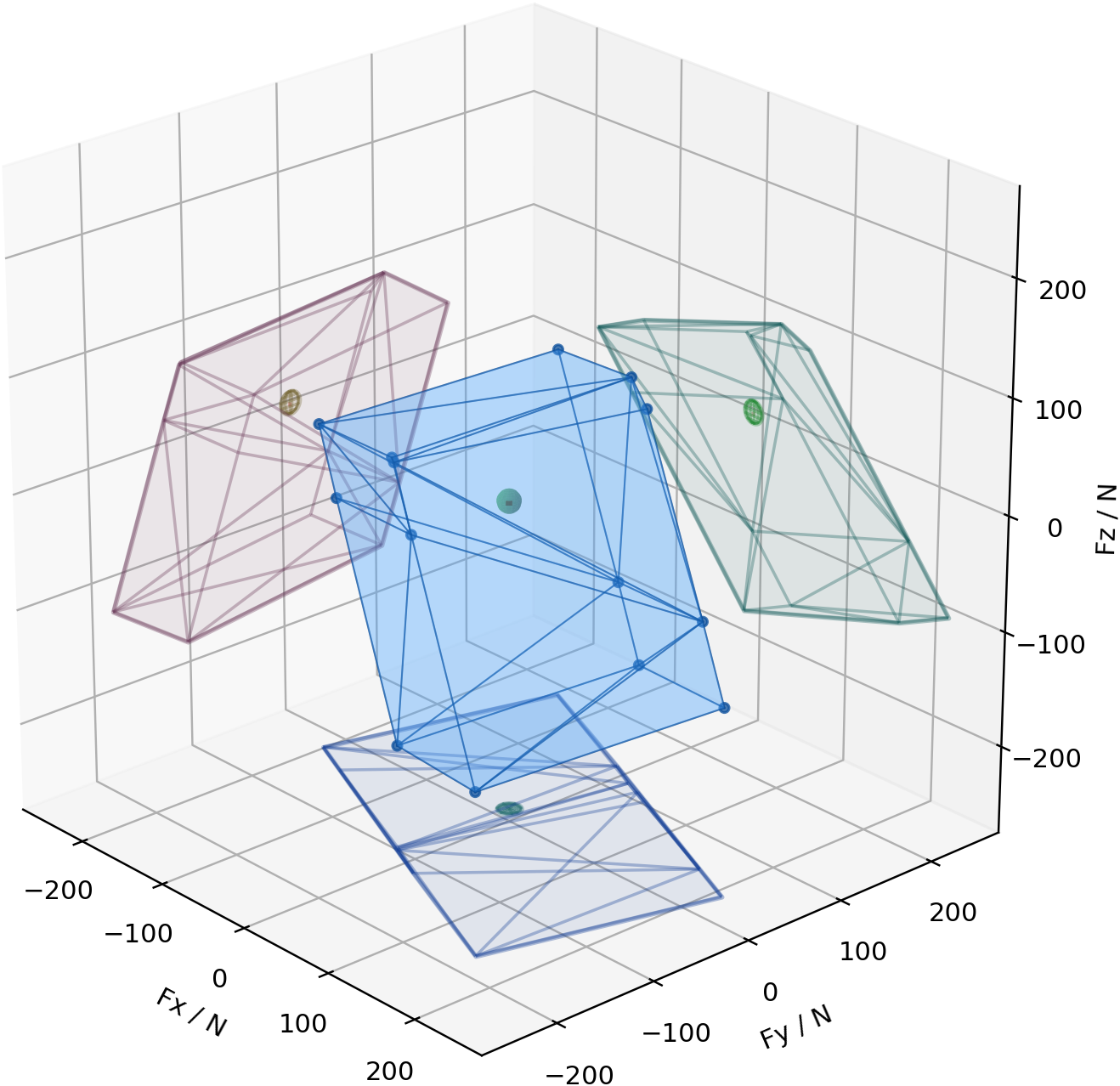}
        &
        \includegraphics[width=0.18\textwidth]{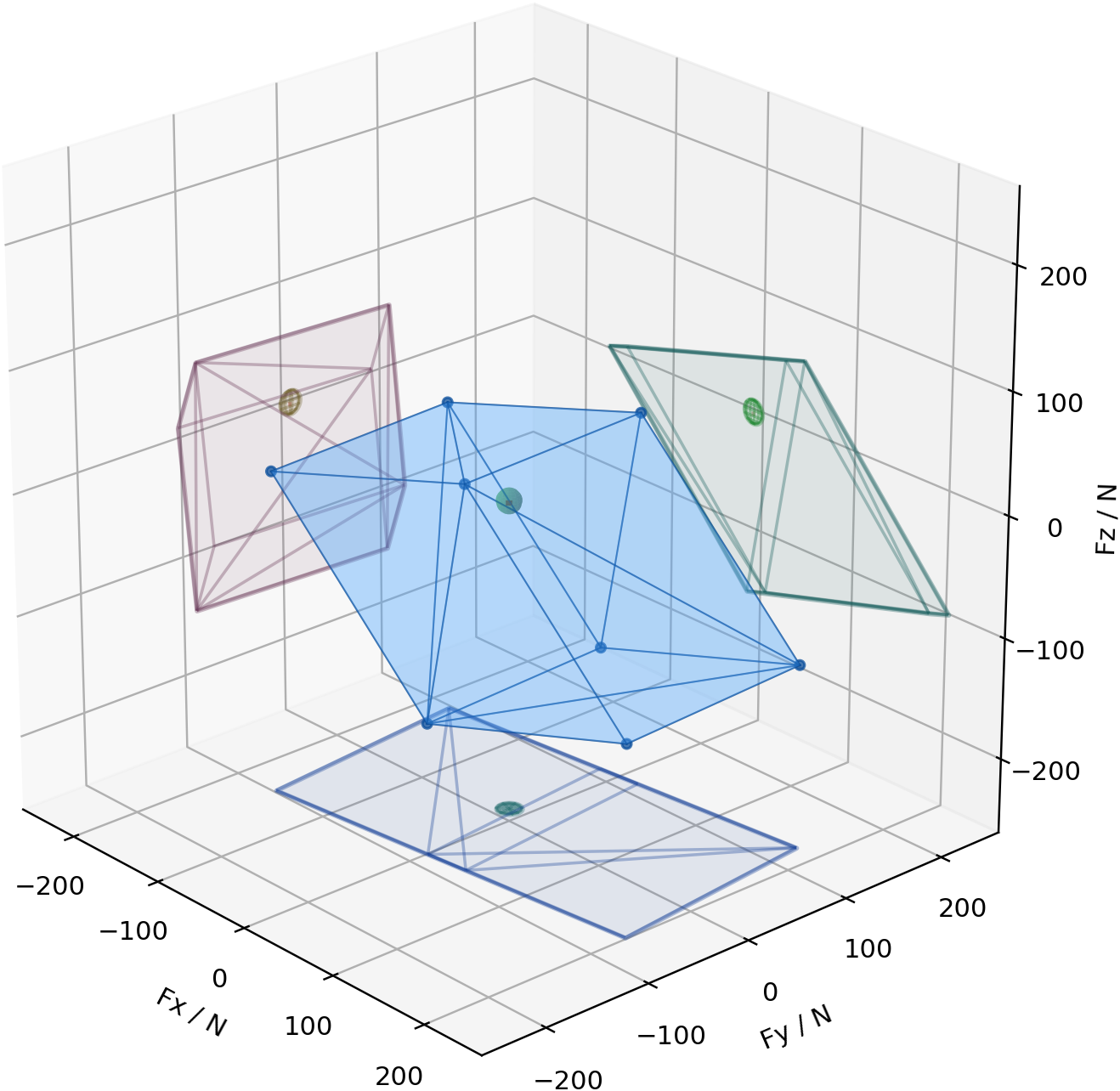}
        &
        \includegraphics[width=0.18\textwidth]{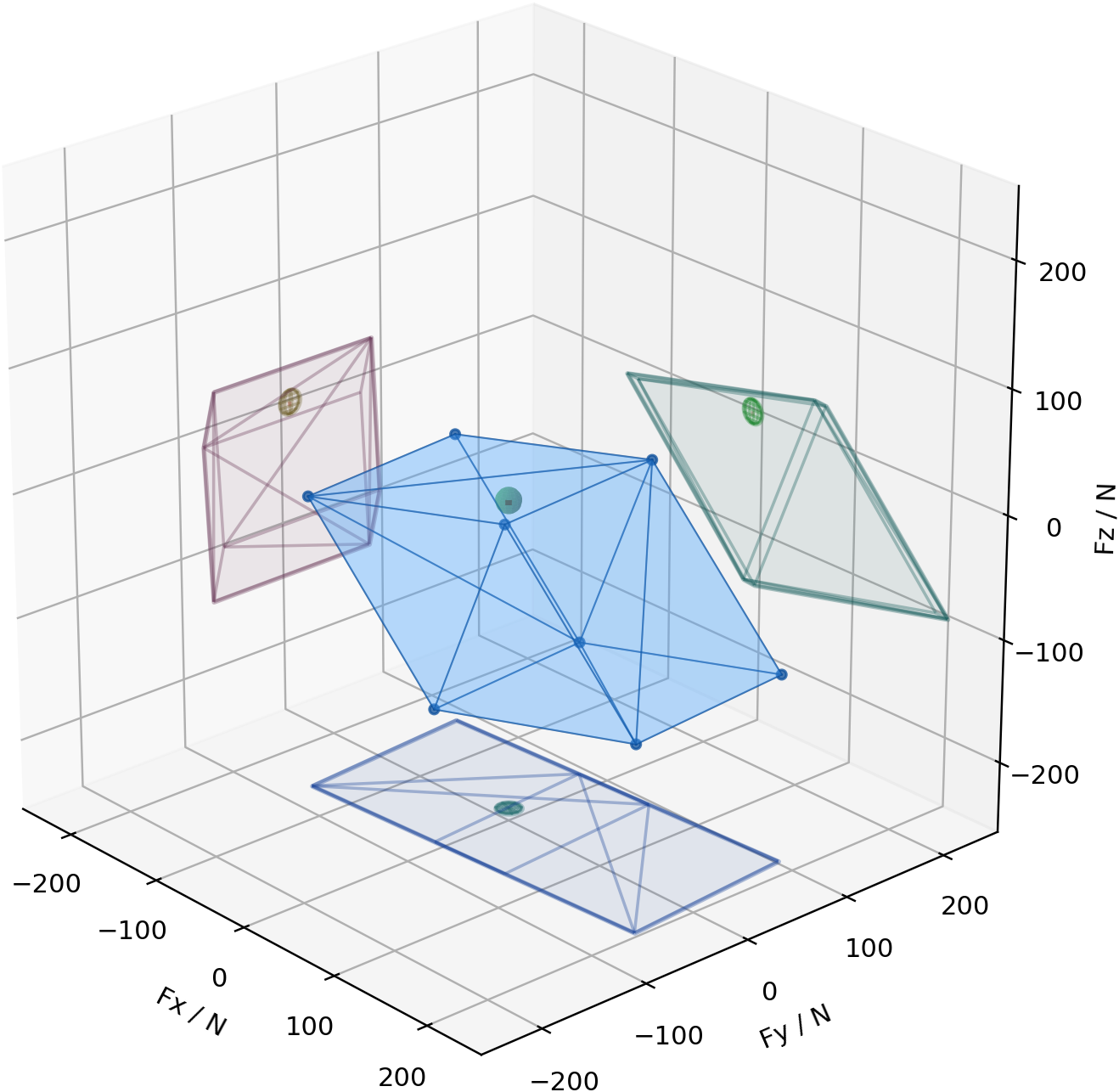}
        \\[-2pt]

        &
        \scriptsize Ours
        &
        \scriptsize Manipulability
        &
        \scriptsize RFP inscribed radius
        &
        \scriptsize RFP Cone
        &
        \scriptsize Teaching point
    \end{tabular}

    \caption{Single-point comparison under different payload conditions. Each column corresponds to one optimization method, and each row corresponds to one beverage load.}
    \label{fig:single_point_2x5_comparison}
\end{figure*}

\begin{figure*}[h]
    \centering

    \includegraphics[width=0.75\textwidth]{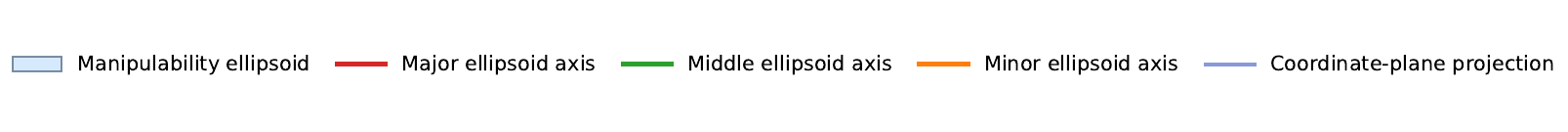}
    \vspace{2pt}

    \setlength{\tabcolsep}{2pt}
    \renewcommand{\arraystretch}{1.0}

    \begin{tabular}{c c c c c c}
        \rotatebox{90}{\scriptsize 2.7 L beverage}
        &
        \includegraphics[width=0.18\textwidth]{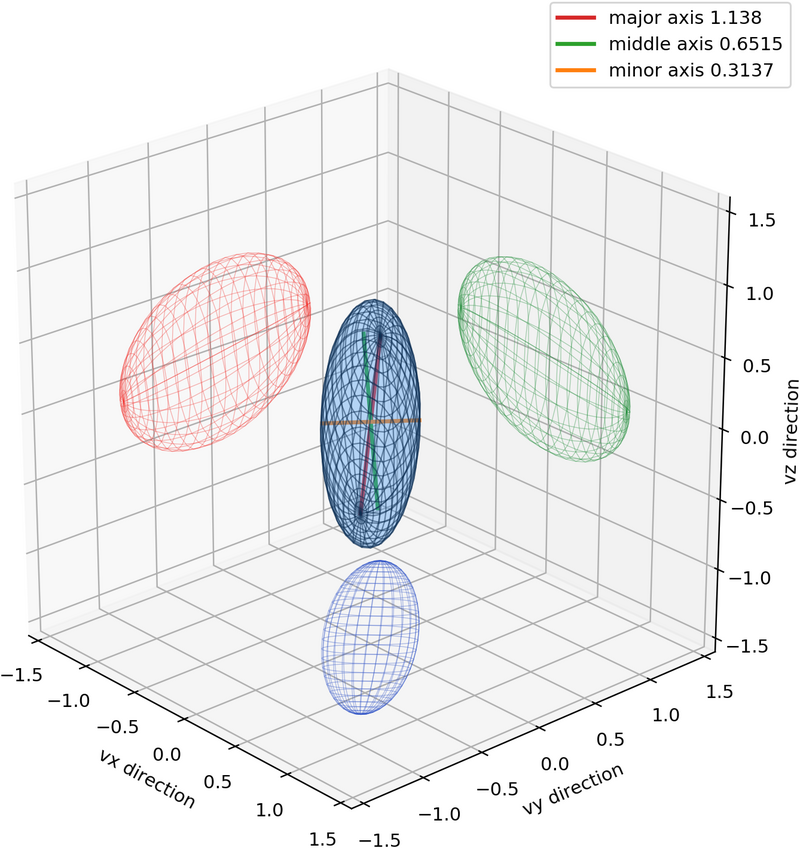}
        &
        \includegraphics[width=0.18\textwidth]{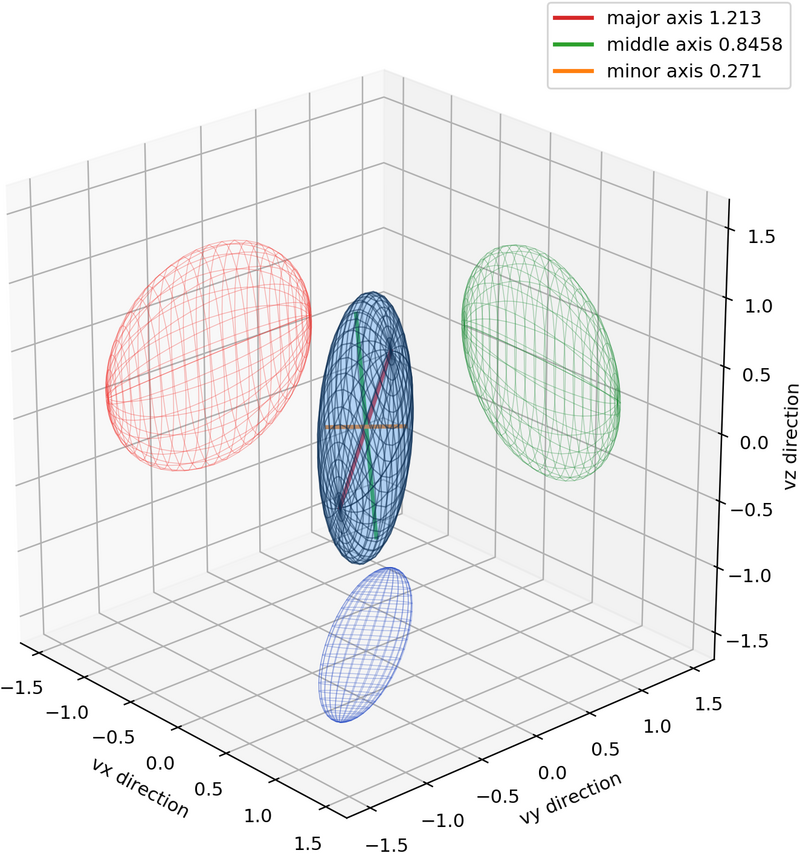}
        &
        \includegraphics[width=0.18\textwidth]{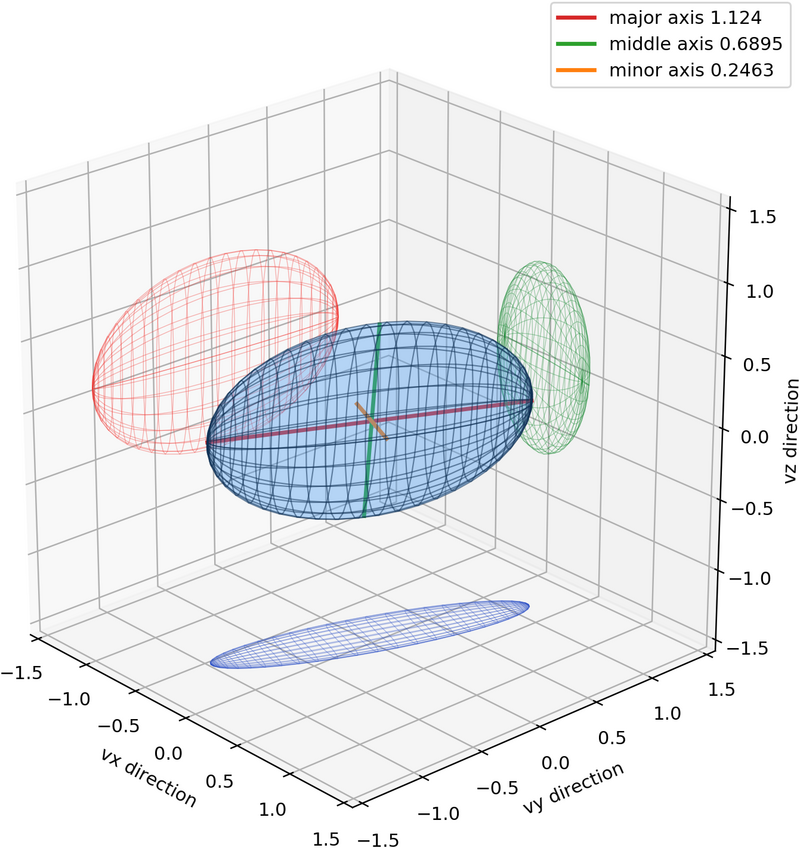}
        &
        \includegraphics[width=0.18\textwidth]{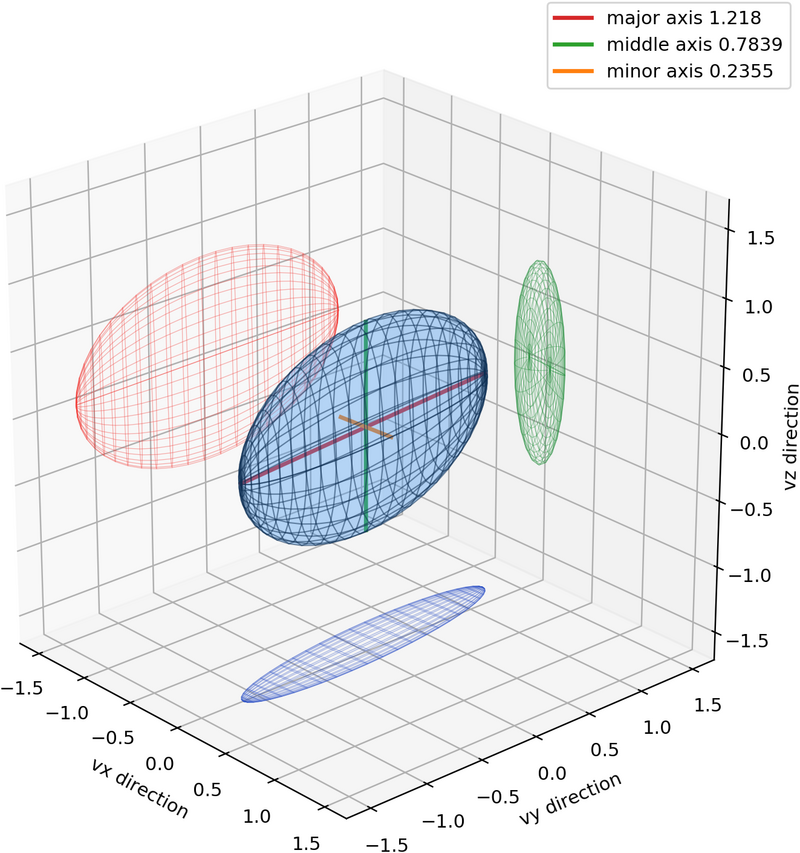}
        &
        \includegraphics[width=0.18\textwidth]{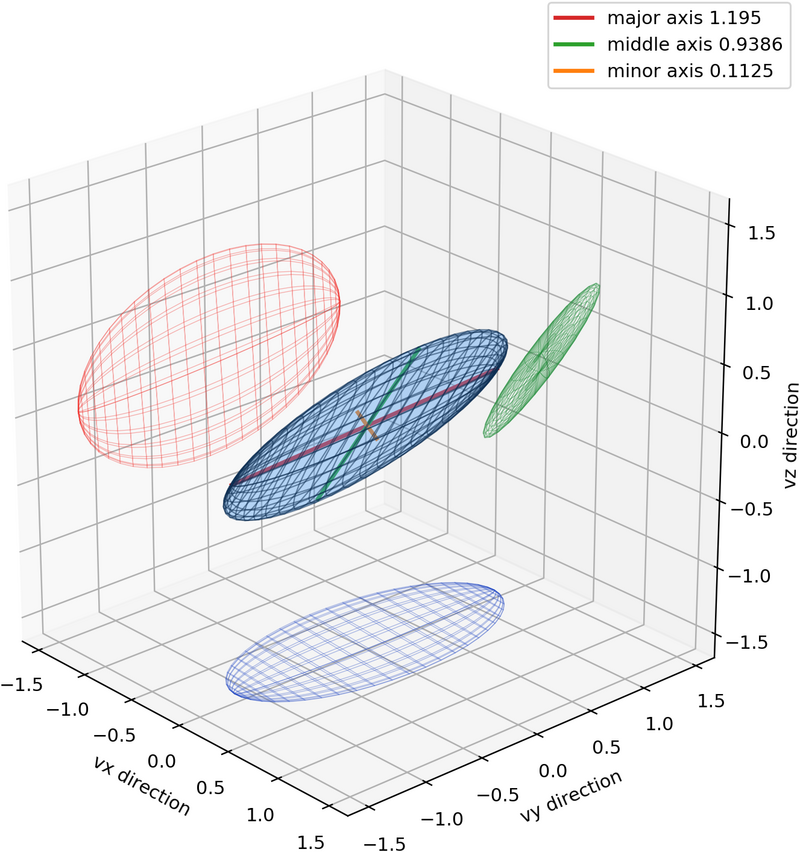}

        \\[-2pt]

        \rotatebox{90}{\scriptsize 0.33 L beverage}
        &
        \includegraphics[width=0.18\textwidth]{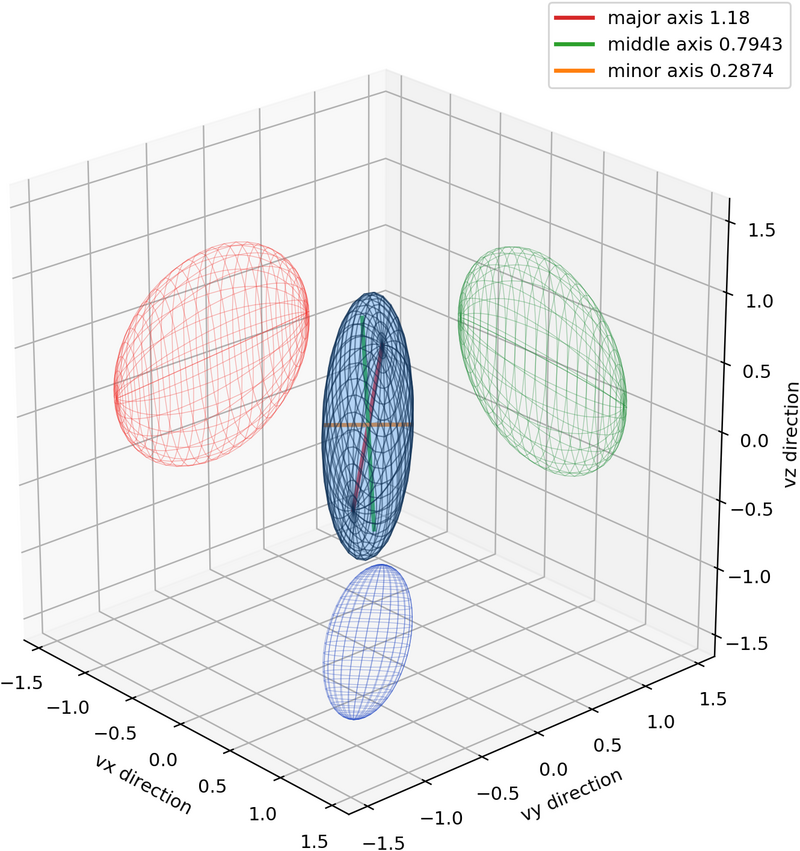}
        &
        \includegraphics[width=0.18\textwidth]{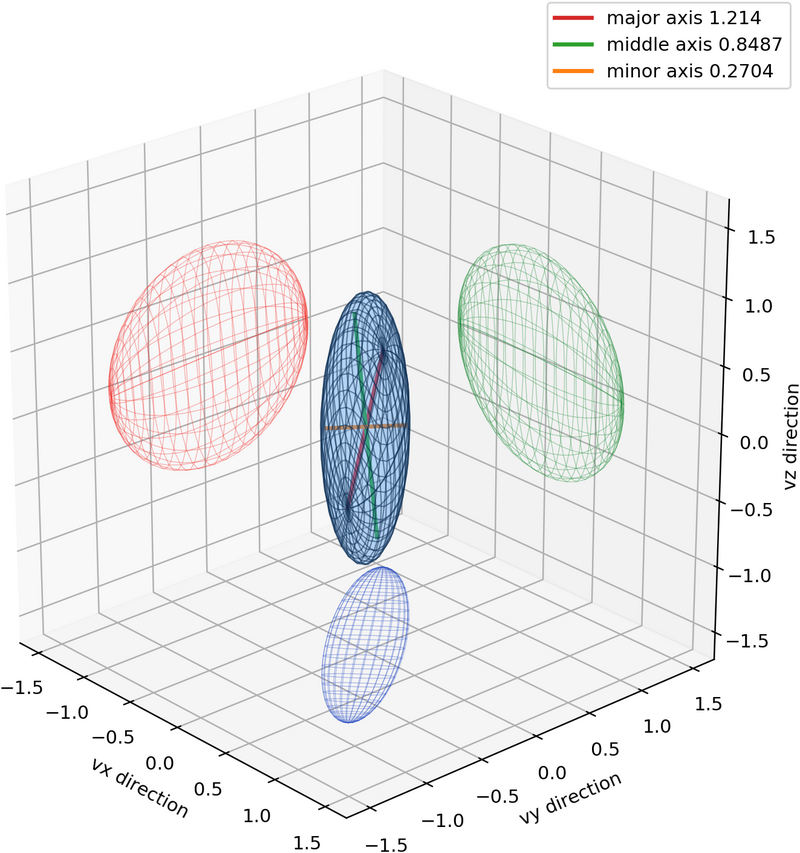}
        &
        \includegraphics[width=0.18\textwidth]{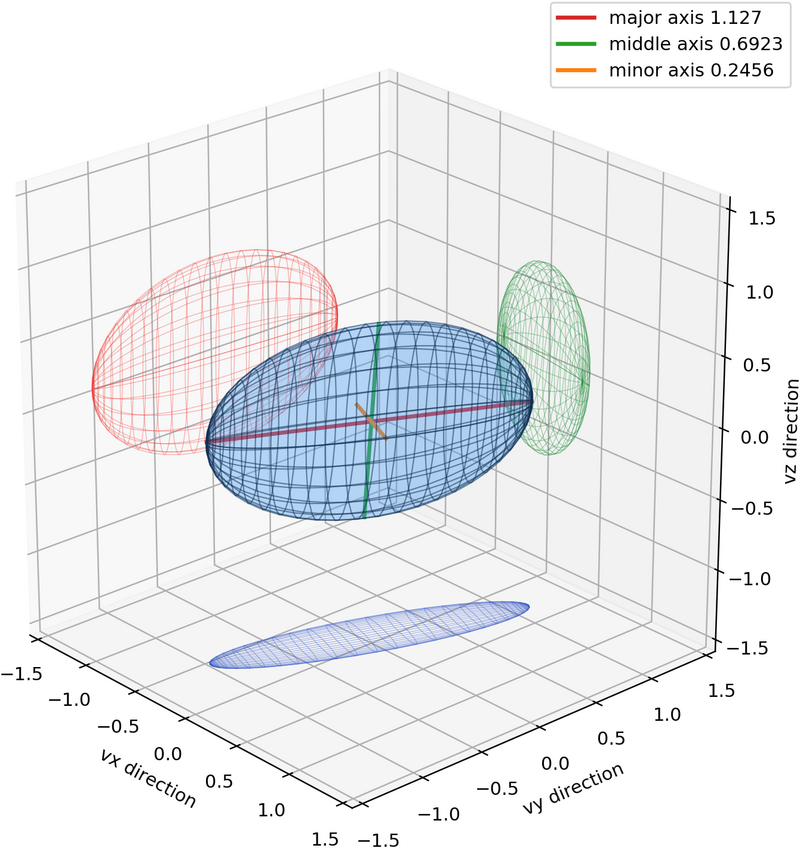}
        &
        \includegraphics[width=0.18\textwidth]{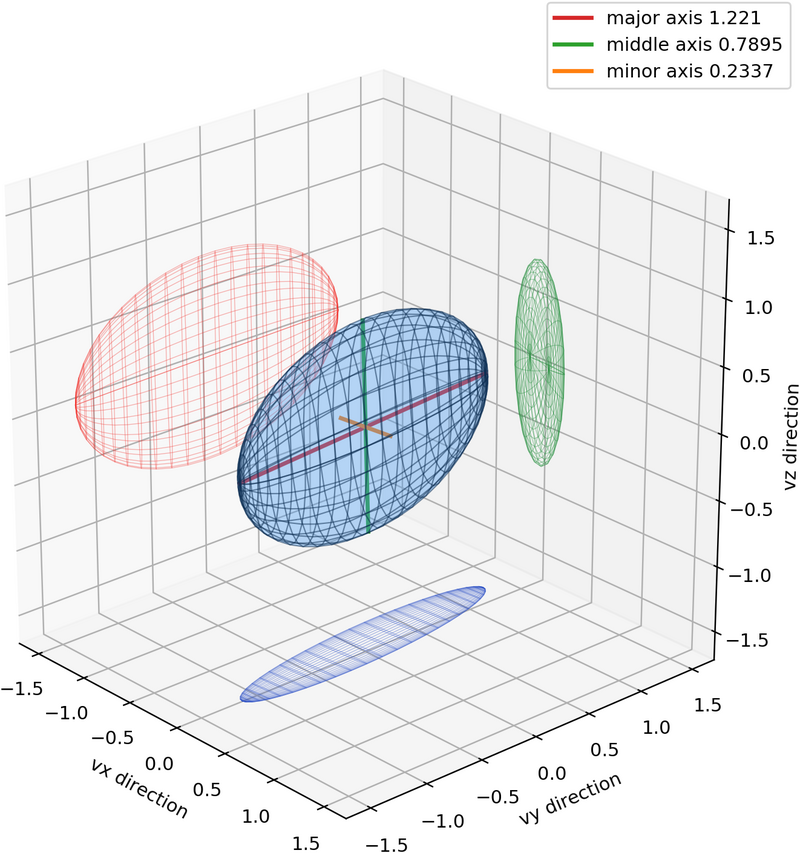}
        &
        \includegraphics[width=0.18\textwidth]{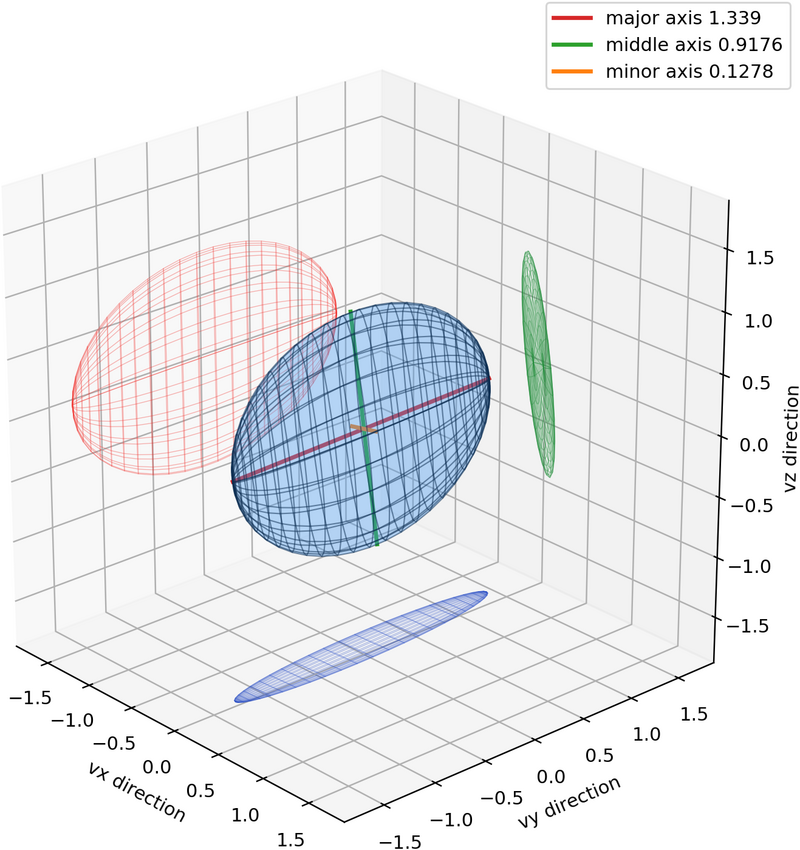}
        \\[-2pt]

        &
        \scriptsize Ours
        &
        \scriptsize Manipulability
        &
        \scriptsize RFP inscribed radius
        &
        \scriptsize RFP Cone
        &
        \scriptsize Teaching point
    \end{tabular}

    \caption{Single-point comparison of manipulability ellipsoids under different payload conditions. Each column corresponds to one optimization method, and each row corresponds to one beverage load.}
    \label{fig:single_point_manipulability_ellipsoid_2x5_comparison}
\end{figure*}

To further evaluate the proposed method in static redundancy resolution, a single-point holding experiment was conducted. Unlike the trajectory experiment, the desired end-effector pose was fixed in this experiment. Therefore, all methods were required to satisfy the same Cartesian pose constraint, while optimizing different joint-space configurations of the mobile manipulator. This setting allows us to directly compare how different optimization objectives redistribute the redundant degrees of freedom and how they affect the force output capability and manipulability at the same task pose. The single-point holding experiment uses the same platform, payload conditions, and compared methods shown in Fig.~\ref{fig:lifting_experiment_overview}, with the desired end-effector pose held fixed.

Fig.~\ref{fig:single_point_realrobot} shows the corresponding real-world robot configurations obtained by different optimization methods under the same end-effector pose constraint. In the following, Fig.~\ref{fig:onepoint} presents the quantitative comparison of directional force capacity and manipulability, whereas Fig.~\ref{fig:single_point_2x5_comparison} and Fig.~\ref{fig:single_point_manipulability_ellipsoid_2x5_comparison} provide geometric visualizations of the residual force polytopes and manipulability ellipsoids for the same set of optimized configurations.

Fig.~\ref{fig:onepoint} compares the directional residual force capacity and manipulability under different payload conditions. When the basket contains the $2.7\,\mathrm{L}$ beverage load, the required vertical supporting force is relatively large. In this case, the proposed method adaptively increases the residual force capability along the task direction according to the estimated payload mass. Its directional force capacity is comparable to that obtained by the RFP inscribed radius method, and remains above the desired force uncertainty range. In contrast, the manipulability-only method does not explicitly account for the required task force. Although it improves the kinematic manipulability, its force capacity falls into the desired force uncertainty range, indicating a risk of insufficient force output. The teaching-point configuration performs even worse in terms of vertical force capability, as its force capacity is lower than the required force, making it difficult for the manipulator to maintain the end-effector pose under the load. This behavior can also be observed in the real-world experiment shown in Fig.~\ref{fig:single_point_realrobot}, where the teaching-point configuration exhibits a noticeably lower arm posture under the 2.7\,L payload, indicating insufficient force capability to support the desired end-effector pose.

When the basket contains only the $0.33\,\mathrm{L}$ beverage load, the required supporting force is much smaller. Under this light-load condition, the proposed method no longer allocates excessive redundancy to increase the force margin. Instead, it automatically reduces the unnecessary force-capability optimization and preserves more redundancy for improving manipulability. As a result, the manipulability achieved by the proposed method is close to that of the manipulability-only method. By contrast, the RFP inscribed radius method still tends to maximize the available force capability, even though the task does not require such a large force margin. This behavior consumes the robot's redundancy to increase force capability that is not needed for the current task.

The residual force polytope visualization in Fig.~\ref{fig:single_point_2x5_comparison} further explains this behavior. The residual force polytope represents the admissible end-effector force set after considering the robot constraints. For the $2.7\,\mathrm{L}$ beverage load, the proposed method, the RFP inscribed radius method, and the RFP Cone method all keep the desired force uncertainty ball inside the residual force polytope, with a visible margin to the polytope boundaries. This indicates that these methods can provide sufficient force capability for holding the payload. However, at the teaching point, the center of the desired force uncertainty ball moves outside the residual force polytope. This means that the robot does not have enough admissible force capability at this configuration to reliably hold the end-effector pose under the heavy payload.

Fig.~\ref{fig:single_point_manipulability_ellipsoid_2x5_comparison} shows the corresponding manipulability ellipsoids. The proposed method produces an ellipsoid shape similar to that of the manipulability-only optimization, especially under the light-load case, where the projected ellipsoid remains relatively balanced across the coordinate planes. In contrast, the two RFP-based comparison methods from~\cite{ferrolhoResidualForcePolytope2021a} tend to sacrifice Cartesian velocity manipulability in order to enlarge the force capability. This can be observed from the thinner projection of their manipulability ellipsoids, especially on the horizontal planes. The RFP Cone method optimizes the configuration mainly according to the required force direction, while the proposed method considers both the direction and magnitude of the required force. Therefore, it can increase the force capability when the load is heavy, and preserve manipulability when the load is light. The teaching-point configuration also shows poor manipulability, with thin projections on both the $xy$ and $xz$ planes.

Overall, the single-point holding experiment demonstrates that the proposed method can adaptively balance force output capability and manipulability according to the task force requirement. It provides sufficient residual force capability for high-load holding tasks, while avoiding unnecessary force-margin maximization in low-load tasks.

\section{Conclusion}

This paper presented a task-oriented force-capability optimization framework that couples perception, force-capability analysis, and whole-body redundancy resolution for mobile manipulators. A VLM estimates the physical properties of the manipulated object and converts them, together with the desired end-effector trajectory, into a desired task-force sequence that captures both gravitational and inertial demands. A force-capability metric, defined as the signed distance between a task-force uncertainty ball and the dynamic residual force polytope, quantifies whether the robot's remaining actuation capacity is sufficient for the task demand along the trajectory. Embedded into a multi-objective whole-body trajectory optimizer together with manipulability, joint-limit avoidance, trajectory smoothness, and base-oscillation suppression, the metric allows the robot to allocate its redundancy adaptively, raising force capability for heavy payloads while recovering manipulability for light ones.

Lifting and single-point-holding experiments on the MOCA platform under $0.33\,\mathrm{L}$ and $2.7\,\mathrm{L}$ payloads confirmed that the proposed method matches dedicated force-maximizing baselines (RFP inscribed radius and RFP cone) under heavy loads, while approaching manipulability-only optimization under light loads, a task-adaptive balance that none of the baselines achieves on its own. Future work includes extending the formulation from translational forces to full six-DoF wrench requirements, incorporating online VLM-based force estimation for real-time replanning, and validating the framework on a broader range of contact-rich tasks such as pushing, drawer opening, and human-robot collaborative transportation.

\section*{Acknowledgment}
This work was supported by the European Union Horizon Project TORNADO under Grant GA 101189557.
This work was supported by the National Key R\&D Program of China under Grant No. 2023YFB4606204.
This work was carried out during Xiao Wang's visit to IIT, with support from the China Scholarship Council.

\section*{CRediT authorship contribution statement}

\textbf{Xiao Wang:} Conceptualization, Methodology, Software,
Investigation, Writing -- original draft, Writing -- review \& editing.

\textbf{Heng Zhang:} Conceptualization, Investigation,
Writing -- original draft, Writing -- review \& editing.

\textbf{Gokhan Solak:} Writing -- review \& editing.

\textbf{Fei Zhao:} Supervision, Resources.

\textbf{Arash Ajoudani:} Supervision, Resources,
Writing -- review \& editing.

\section*{Declaration of generative AI and AI-assisted technologies in the manuscript preparation process}

The authors used ChatGPT to assist with language editing and checks of technical wording and equations.
All suggestions were reviewed and revised by the authors, who take full responsibility for the final manuscript.
\section*{Declaration of competing interest}

The authors declare no competing financial or non-financial interests related to this work.

\section*{Data availability}
The experimental data supporting this study are available
in Mendeley Data at \url{https://doi.org/10.17632/ykhftpyvkg.1}.

\bibliographystyle{elsarticle-num}
\bibliography{ref}

\end{document}